%% file: main.tex
\documentclass[3p, preprint,11pt, review]{elsarticle}

\usepackage{hyperref}
\usepackage{enumitem}
\usepackage{amsmath}
\usepackage{amsfonts}
\usepackage{amsthm}
\usepackage{algorithm}
\usepackage{setspace}
\usepackage{tabularx}
\usepackage{xltabular}
\usepackage[super]{nth}
\usepackage{booktabs}
\usepackage{subcaption}
\usepackage{algpseudocode}

\algnewcommand{\LineComment}[1]{\State \(\triangleright\) #1}
\usepackage[acronym]{glossaries}
\newacronym{amr}{AMR}{Autonomous Mobile Robot}
\newacronym{drl}{DRL}{Deep Reinforcement Learning}
\newacronym{rl}{RL}{Reinforcement Learning}
\newacronym{momdp}{MOMDP}{Multi-Objective Markov Decision Process}
\newacronym{ppo}{PPO}{Proximal Policy Optimization}
\newacronym{gin}{GIN}{Graph Isomorphism Network}
\newacronym{gcn}{GCN}{Graph Convolutional Network}
\newacronym{milp}{MILP}{Mixed-Integer Linear Programming}
\newacronym{agv}{AGV}{Autonomous Guided Vehicle}

\newcommand{\nop}[1]{}

\journal{Elsevier}

\biboptions{authoryear}

\begin{document}

\begin{frontmatter}
\title{Learning Efficient and Fair Policies for Uncertainty-Aware Collaborative Human-Robot Order Picking}

\author[inst1,inst2]{Igor G. Smit\corref{cor1}}
\author[inst1]{Zaharah Bukhsh}
\author[inst2]{Mykola Pechenizkiy}
\author[inst3]{Kostas Alogariastos}
\author[inst3]{Kasper Hendriks}
\author[inst1]{Yingqian Zhang}
\cortext[cor1]{Corresponding author. \textit{E-mail address:} i.g.smit@tue.nl}
\address[inst1]{Department of Industrial Engineering \& Innovation Sciences, Eindhoven University of Technology, P.O. Box 513,
5600 MB Eindhoven, the Netherlands}
\address[inst2]{Department of Mathematics \& Computer Science, Eindhoven University of Technology, P.O. Box 513,
5600 MB Eindhoven, the Netherlands}
\address[inst3]{Vanderlande Industries B.V., P.O. Box 18, 5460 AA Veghel, the Netherlands}
\begin{abstract}
In collaborative human-robot order picking systems, human pickers and Autonomous Mobile Robots (AMRs) travel independently through a warehouse and meet at pick locations where pickers load items onto the AMRs. 
In this paper, we consider an optimization problem in such systems where we allocate pickers to AMRs in a stochastic environment. We propose a novel multi-objective Deep Reinforcement Learning (DRL) approach to learn effective allocation policies to maximize pick efficiency while also aiming to improve workload fairness amongst human pickers. 
In our approach, we model the warehouse states using a graph, and define a neural network architecture
that captures regional information and effectively extracts representations related to efficiency and workload. 
We develop a discrete-event simulation model, which we use to train and evaluate the proposed DRL approach. 
In the experiments, we demonstrate that our approach can find non-dominated policy sets that outline good trade-offs between fairness and efficiency objectives.
The trained policies outperform the benchmarks in terms of both efficiency and fairness. Moreover, they show good transferability properties when tested on scenarios with different warehouse sizes.  The implementation  of the simulation model, proposed approach, and experiments are published. 
\end{abstract}

\end{frontmatter}
\section{Introduction}

Order picking is one of the most fundamental and costly processes in logistics. In conventional warehouses, human pickers typically spend up to 90\% of time on picking activities, and 55\% of all operating costs are commonly attributed to order picking \citep{Dukic2007}. To improve picking efficiency, collaborative human-robot order picking has recently gained increasing attention, where human pickers and \glspl{amr} travel independently 
and meet at pick locations, where pickers load retrieved items onto the \glspl{amr}.

While optimizing \emph{efficiency} (total picking time) is a dominant focus within both traditional \citep{koster2007} and robotized warehousing settings \citep{Azadeh2019}, our study also takes into account the \emph{workload fairness}, an objective often ignored in the literature. Existing solutions \citep{Zulj2022, Srinivas2022, Loffler2022} typically focus on deterministic scenarios
and optimizing for efficiency. However, the sole focus on efficiency can negatively impact human well-being. 
If some pickers must pick much larger/heavier workloads than others, it can place considerable physical and mental strain on them. This increases injury risk, reduces picker well-being and work satisfaction, and may violate ergonomic guidelines. In addition, the deterministic nature of existing methods drastically limits their applicability. Collaborative picking involves many pickers and \glspl{amr} interacting in complex ways. These systems have many sources of uncertainty, such as random movement speeds, uncertain pick times, congestion, and other disruptions. These uncertainties can drastically affect the process and make predetermined schedules infeasible. Hence, there is a need for an online stochastic solution approach.

To address these challenges, we propose 
 a novel multi-objective \gls{drl} approach that learns allocation policies to jointly optimize efficiency and fairness. As decision-makers may value these objectives differently,  we explicitly outline the achievable trade-offs. Our approach frames the warehouse setting as an online decision-making problem that includes uncertainty and disruptions in dynamic environments. 
We model the warehouse states as a graph as this allows for a representation that can easily adjust to different warehouse instances. As existing graph neural networks do not effectively handle the long-range dependencies in warehouse settings, we propose 
a novel 
Aisle-Embedding Multi-Objective Aware Network (AEMO-Net)
architecture to effectively capture regional information in warehouses. In addition, we show a feature separation principle to effectively extract workload and efficiency information in multi-objective \gls{drl}. 

We conduct extensive experiments, 
and show that we can find non-dominated policy sets that outline good trade-offs between fairness and efficiency. Compared to greedy and business rule benchmarks, the multi-objective policy sets contain multiple policies that achieve superior performance on both objectives. 
For large warehouses, we achieve policies improving efficiency by 20\% while reducing unfairness (workload standard deviation) by 90\% compared to the company benchmark. Moreover, the trained policies generalize and transfer to new configurations like different picker/\gls{amr} numbers and warehouse sizes, while maintaining better performance over benchmarks. 
 
Our work offers the following key contributions: (1) We have a unique problem setting of human-robot collaborative order picking with a focus on the workload fairness of human pickers. This is the first study that
optimizes both system performance and workload fairness, while we also handle a highly stochastic environment directly derived from a real-world practical use case;  
(2) We develop a multi-objective \gls{drl} approach that explicitly generates a non-dominated set of policies outlining the trade-offs between efficiency and fairness with different underlying metrics. Experiments demonstrate its good performance and transferability;  
(3) We propose a lightweight graph-based neural network architecture, which can efficiently capture spatial information in warehouses in a less computationally demanding but more expressive way than standard graph neural networks. It also effectively handles multiple feature groups, which contain information related to different objectives.
\par
The paper is structured as follows: Section 2 provides a review of the literature, Section 3 is devoted to a formal definition of the problem, and Section 4  introduces our DRL-Guided Picker Optimization methodology. Detailed descriptions of the experimental setup are presented in Section 5. Finally, we synthesize our findings and discuss the implications of our work in Section 6. 

\section{Related Work}
In this section, we first outline the existing collaborative picking works. Then, we show the promise of \gls{drl} for online optimization problems and briefly describe current works addressing fairness in \gls{drl}. 
\subsection{Collaborative Picking}
Most existing works in collaborative picking aim at 
optimizing warehouse layouts \citep{Lee2019} and zoning strategies \citep{Azadeh2020}.
Recently, several studies have targeted  
operations optimization, with focus on efficient picker routing 
\citep{Loffler2022}, and 
tardiness minimization 
\citep{Zulj2022} using \gls{milp} models. \citet{Loffler2022} studied picker routing in \gls{agv}-assisted picker-to-parts order picking in single-block, parallel-aisle warehouses. They developed an exact polynomial time algorithm to minimize the total traveled distance for cases with fixed picking sequences. They showed good performance compared to traditional order picking. Similarly, \citet{Zulj2022} considered a different variant in which \glspl{amr} collect items and transfer them to the central warehouse depot. They use a heuristic that solves an \gls{milp} formulation considering order batching and sequencing. 
\citet{Srinivas2022} also focused on minimizing tardiness. They considered a problem most similar to ours in which both pickers and \glspl{amr} can freely move through the warehouse. They integrated order batching, sequencing, and picker-robot routing in their method. Again, their method used an \gls{milp} model. They proposed a restarted simulated annealing approach with adaptive neighborhood search improving exploration and exploitation. They showed near-optimal results for several problem instances.
Lastly, \citet{Xie2022} proposed two \gls{milp} formulations and a variable neighborhood search heuristic for a zone-based collaborative picking system. 

\subsection{Deep Reinforcement Learning for Online Planning}
Thus, most existing methods consider deterministic scenarios that create full solutions in advance, ignoring the fact that processes in warehouses are highly stochastic due to disruptions and uncertainties. \citet{Beeks2022} and \citet{Cals2021} have shown the advantages of \gls{drl} approaches in tackling uncertainties in warehouses. However, they focus on a very different optimization problem, i.e., order batching, and like other existing methods,  
do not incorporate human factors.   
Regarding \gls{drl} for allocation or matching problems, 
\citet{Alomrani2021} solve an online bipartite matching problem in which a fixed entity set must be matched with incoming entities.
Our problem can also be considered as a matching problem but has additional complexity such as spatial relations, interdependent availability of nodes, both items/\glspl{amr} and pickers being uncertain over time, and fairness as an additional optimization objective. 
\par
Recently, \citet{Begnardi2023} proposed a \gls{drl} approach to solving a similar matching problem as ours related to collaborative order picking. However, they study a much more simplified environment. For instance, they do not regard a realistic warehouse layout, do not explicitly incorporate \gls{amr} interactions and congestion, and have limited stochasticity. In addition, they do not consider workload fairness. 
\subsection{Fairness in Reinforcement Learning}
\citet{Gajane2022} survey 
a variety of case-specific studies on fair RL, e.g.\ traffic light control \citep{Li2020, Raeis2021}, UAV control \citep{ 
Qi2020, 
Nemer2022}, or resource allocation \citep{Zhu2018}. These studies consider single-objective \gls{drl} with handcrafted reward functions to find one specific policy and do not apply multi-objective \gls{drl} to find trade-offs. 
\citet{Siddique2020} propose a 
method to learn fair policies in \gls{drl}. They used the generalized Gini function to determine a fair policy over multiple agents and modeled the problem as a \gls{momdp}. 
The difference between their multi-objective modeling and ours is that we have two optimization objectives, efficiency and workload fairness, while their performance objective is solely on optimizing a fair reward distribution among individuals. Note that in cases where performance efficiency is vital, like most operations optimization problems, the approach by \citet{Siddique2020} for individual fairness is not applicable.  

\section{Problem Formulation}
We consider a traditional warehouse layout with vertical, parallel aisles with storage racks on both sides of the aisles. At the top and the bottom, two horizontal cross-aisles connect the vertical aisles. The \glspl{amr} traverse vertical aisles unidirectionally.
Human pickers can move in both directions in each aisle. 
Both pickers and \glspl{amr} can move in either direction within the horizontal cross-aisles. 
We address an optimization problem where human pickers are assigned to AMRs (or items) in a collaborative picking environment. 
Each AMR is assigned with a set of ``pickruns", specifying the items to pick, corresponding locations, and the required collection sequence. 
The \gls{amr} moves toward its first pick destination. Upon arrival, 
it waits until a human picker arrives and places the required items on it. Then, the \gls{amr} proceeds to its next destination. 
This process continues until the \gls{amr} completes its entire pickrun. 
After unloading at a drop-off location, the \gls{amr} receives a new pickrun, and the cycle restarts. This happens for many \glspl{amr} simultaneously. 
The human pickers are distributed through the warehouse. When idle, a picker requests and receives a new   picking location where 
the picker retrieves the items from the shelves and loads them onto the \glspl{amr}. 

The standard optimization objective in warehousing operations is to maximize efficiency. In our case, we minimize the total time to complete the set of pickruns. In addition, we consider workload fairness in optimization. 
Most warehouses contain diverse product assortments of varying weights. For instance, in supermarket warehouses (our study case), 
some products weigh just a kilogram, like boxes with crisps, while others can be ten or fifteen times as heavy, such as packs with drinks. 
Therefore,
we measure the picker workload by the total mass of the lifted products they must pick, and measure the fairness by the standard deviation of the workloads of all pickers.  
 
To illustrate the problem formally, we formulate the deterministic version of the optimization problem as an \gls{milp}.  
In this formulation, we assume that each \gls{amr} only fulfills one pickrun for simplicity. 
Let $i \in \mathcal{N}$ denote the set of all items/orders that must be picked, $r \in \mathcal{R}$ a set of \glspl{amr}, and $k \in \mathcal{K}$ the human pickers. 
We have two binary decision variables: $A_{i,k}$ and  $U_{i,i'}$, where $A_{i,k}=1$ if picker $k$ is assigned to item $i$,and $U_{i,i'}=1$ if $i$ must be retrieved before item $i'$ by the same picker, 0 otherwise. The list of  variables can be found in Table \ref{tab:notation}. We define the problem as follows.
\begin{table}[tb]
\setlength{\tabcolsep}{4.1pt}
\centering
\begin{footnotesize}
\begin{tabular}{ll}
 \toprule
 Variables & Descriptions \\
 \midrule
   $A_{i,k}$ & Decision variable; 1 if picker $k$ is assigned to item $i$, 0 otherwise.\\
    $U_{i,i'}$ & Decision variable; 1 if item $i$ must be picked before $i'$ by same picker.\\
    $B_{i,k}^K$ & Time when picker $k$ arrives at item $i$.\\
    $F_{i,k}^K$ & Time when picker $k$ is ready to leave item $i$'s location.\\
    $B_{i,r}^R$ & Time when item $i$ can be placed on \gls{amr} $r$.\\
    $F_{i,r}^R$ & Time when item $i$ has been placed on \gls{amr} $r$.\\
    $C_{r}^R$ & Completion time of the pickrun of \gls{amr} $r$.\\
    $C$ & Completion time of the last pickrun.\\
    $W_k$ & The total workload of picker $k$. \\
    $M$ & Sufficiently large positive number.\\
    $\tau_{i,i'}^K$ & Travel time from item $i$ to  item $i'$ by human pickers.\\
    $\tau_{i,i'}^R$ & Travel time from item $i$ to item $i'$ by \glspl{amr}.\\
    $\tau_{r, i}^{o,R}$ &  Travel time from starting location of \gls{amr} $r$ to location $i$.\\
    $\tau_{k, i}^{o,K}$ & Travel time from starting location of picker $k$ to location $i$.\\
    $\eta_i^L$ & Time to place item $i$ on an \gls{amr}.\\
    $u_{i,i'}^R$ & 1 if \gls{amr} $r$ collects  $i$ before $i'$, 0 otherwise.\\
    $a_{i,r}^R$ & 1 if \gls{amr} $r$ must transport item $i$, 0 otherwise.\\
    $w_i$ & The workload value of item $i$.
  \\ \bottomrule
\end{tabular}
\caption{The variables of the picker allocation problem.\label{tab:notation}
}
\end{footnotesize}
\end{table}
\[
\min C,  ~~~~ \text{and} ~~~~ \min \sigma(W_1, W_2,\dots, W_{|\mathcal{K}|}), 
\]
\noindent subject to
 \begingroup
\allowdisplaybreaks
\begin{align}
   &\sum_{k\in\mathcal{K}} A_{i,k} = 1 \quad  &&\forall i\in \mathcal{N} \\
    &A_{i,k} - A_{i',k} \leq 1 - (U_{i,i'} + U_{i',i}) \quad &&\forall i,i'\in\mathcal{N}, i \neq i',k\in\mathcal{K} \\
    &A_{i,k} + A_{i',k} \leq 1 + (U_{i,i'} + U_{i',i}) \quad &&\forall i,i'\in\mathcal{N}, i \neq i',k\in\mathcal{K} \\
    &B_{i.k}^K \geq \tau_{k, i}^{o,K} - M \cdot (1-A_{i,k}) \quad &&\forall i \in \mathcal{N}, k \in \mathcal{K} \\
    &
    \begin{aligned}
    \sum_{k\in\mathcal{K}}B_{i.k}^K \geq &\sum_{k\in\mathcal{K}}F_{i',k}^K + \tau_{i',i}^K \cdot U_{i',i}\\ &- M \cdot (1-U_{i',i}) \quad
    \end{aligned}
    && \forall i,i'\in \mathcal{N},  i \neq i'  \\
    & F_{i,k}^K \geq F_{i,r}^R -M \cdot \left(2-A_{i,k} - a_{i,r}^R\right) \quad && \forall i\in\mathcal{N}, k \in \mathcal{K}, r\in\mathcal{R} \\
    & \sum_{r \in \mathcal{R}} B_{i,r}^R \geq \left(\sum_{r \in \mathcal{R}} F_{i',r}^R + \tau_{i',i}^R \right) \cdot u_{i',i}^R \quad && \forall  i, i' \in \mathcal{N}, i \neq i'  \\
    & B_{i,r}^R \geq \tau_{r, i}^{o,R} \cdot a_{i,r}^R \quad && \forall i \in \mathcal{N}, r \in \mathcal{R}  \\
    & B_{i,r}^R \geq B_{i,k}^K - M \cdot (2 - A_{i,k} - a_{i,r}^R) \quad && \forall i \in \mathcal{N}, r \in \mathcal{R}, k \in \mathcal{K}  \\
    &F_{i,r}^R=B_{i,r}^R + \eta_{i}^L\cdot a_{i,r}^R \quad && \forall i\in\mathcal{N},r\in\mathcal{R}, \\
    &C_r^R \geq F_{i,r}^R \quad && \forall i \in \mathcal{N}, r \in \mathcal{R} \\
    & C \geq C_r^R \quad && \forall r \in \mathcal{R}  \\
    & W_k = \sum_{i\in\mathcal{N}} w_i \cdot A_{i,k} \quad &&\forall k \in\mathcal{K}  \\
    & B_{i,k}^K \leq M \cdot A_{i,k} \quad && \forall i \in \mathcal{N}, k\in\mathcal{K}  \\
    &F_{i,k}^K \leq M \cdot A_{i,k} \quad && \forall i \in \mathcal{N}, k\in\mathcal{K}  \\
    &B_{i,r}^R \leq M \cdot a_{i,r}^R &&\forall i\in\mathcal{N},r\in\mathcal{R} \\
    &F_{i,r}^R\leq M\cdot a_{i,r}^R \quad &&\forall i\in\mathcal{N}, r\in\mathcal{R}  \\
    & A_{i,k}, U_{i,i'} \in \{0,1\} \quad &&
    \begin{aligned}
    \forall &i,i'\in\mathcal{N}, k\in\mathcal{K}, r \in \mathcal{R}
    \end{aligned}  \\
    &B_{i,k}^K, F_{i,k}^K, F_{i,r}^R, B_{i,r}^R, C_r^R, C \geq 0 \quad &&
    \begin{aligned}
    \forall &i\in\mathcal{N}, k\in\mathcal{K}, r\in\mathcal{R}
    \end{aligned}
\end{align}
\endgroup

The first constraint ensures each item is picked by just one picker. Constraints 2-3 define the relative order of two items picked by the same picker. 
Constraints 4-5 compute the  
time when a picker can pick an item, and Constraint 6 indicates when a picker can leave a location. 
Constraints 7-9
describe when an \gls{amr} is ready for an item to be placed on it. 
Then, an \gls{amr} has been loaded and can leave the location after the associated pick time has passed (\nth{10} constraint).
Equation 11 bounds the completion time of an \gls{amr} pickrun to the time at which the \gls{amr} has finished its last pick. Constraint 12 computes the efficiency objective value. In Constraint 13, the total workload of a picker is the sum of the workloads of each pick. 
Constraints 14-15 ensure the beginning and finishing times of picker actions related to an item are only set when the picker is assigned to pick this item. Constraints 16-17 enforce this for \glspl{amr}. The last two constraints specify the decision variables $A_{i,k}$ and $U_{i, i'}$ are binary and the time-related variables are non-negative.

We focus on optimizing the picker-AMR allocation decisions, and assume the releasing strategies of the orders and the \gls{amr} routing are fixed. Despite these fixed strategies and pre-determined pickruns, the environment is highly stochastic in reality. 
Therefore, deterministic optimization methods are not preferred.
Instead, we model the problem as a sequential decision-making problem and develop a \gls{drl} approach to learn good allocation policies that account for inherent stochasticity. 

\section{DRL-Guided Picker Optimization}
\begin{figure}[ht]
    \centering
    \includegraphics[width=0.65\linewidth]{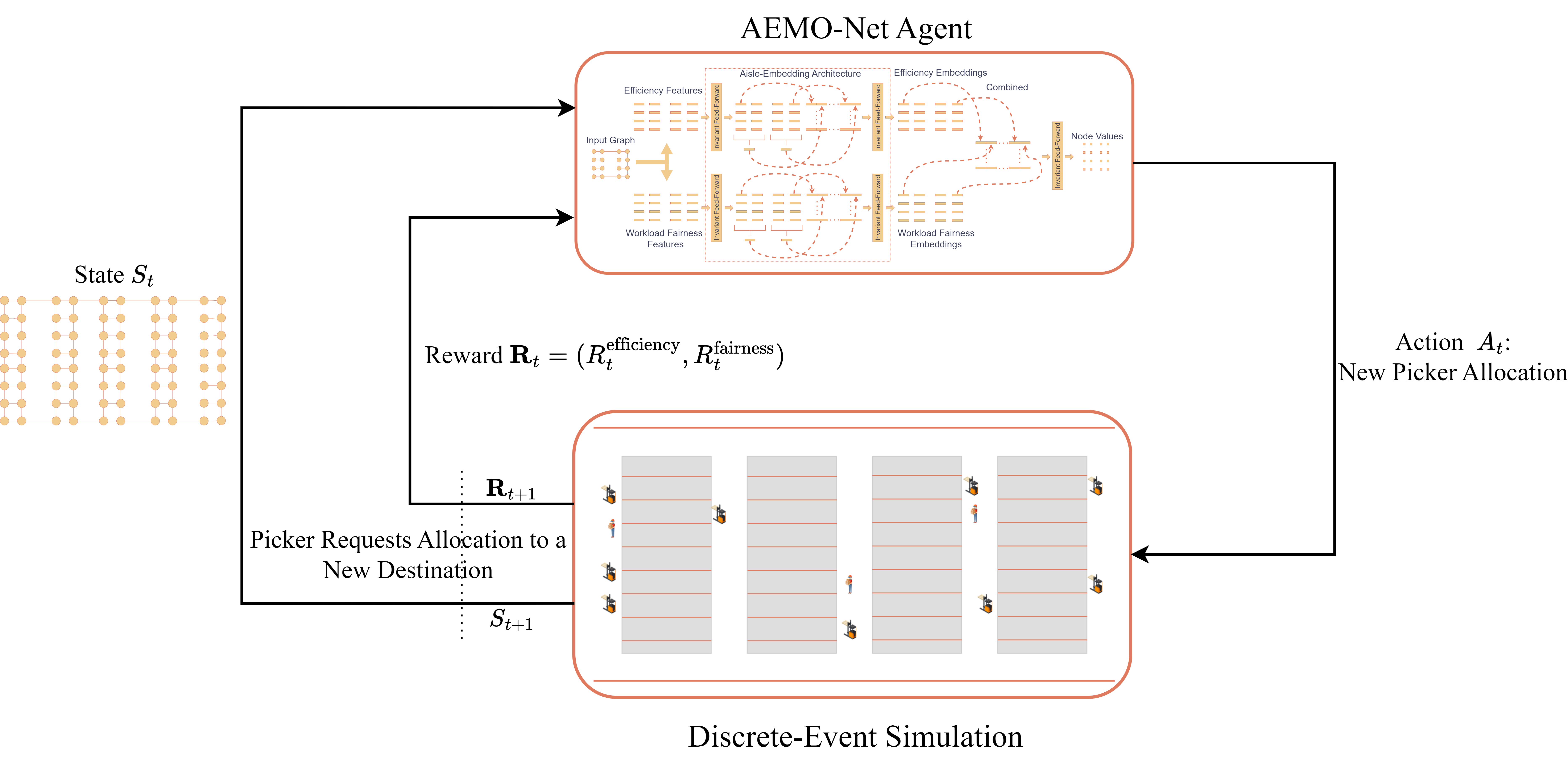}
    \caption{Overview of DRL-Guided Picker Optimization.}
    \label{fig: RL_overview}
\end{figure}

To address the problem, we propose the \gls{drl}-Guided Picker Optimization approach\footnote{cf.~\url{https://github.com/ai-for-decision-making-tue/DRL-Guided-Picker-Optimization}}. Figure \ref{fig: RL_overview} offers an overview of this method. The approach builds upon several key components, which we further elaborate on in this section. Firstly, we develop a discrete-event simulation model representing the collaborative picking system, oultined in Section \ref{sec:simulation_model}. Secondly, in Section \ref{sec:mdp}, we formalize the \gls{momdp} which provides the general framework in which the \gls{drl} agent can interact. Thirdly, we propose a novel neural network architecture in Section \ref{sec:aemo_net}. And lastly, we introduce the learning algorithm in Section \ref{sec:learning_algorithm}. 

\subsection{Simulation Model} \label{sec:simulation_model}
We develop a discrete-event simulation model wrapped within the OpenAI Gym \citep{Brockman2016} interface to represent the collaborative picking system. We use product and order picking data from a real-world grocery distribution center. 
Several sources of randomness are modeled to simulate a stochastic environment: pick times, picker and \gls{amr} speeds, picking disruption occurrences and duration, and \gls{amr} overtaking delays. We apply our \gls{drl} framework within this simulation environment, which we describe in more detail below.
\par
\begin{figure}
    \centering
    \includegraphics[width=0.65\linewidth]{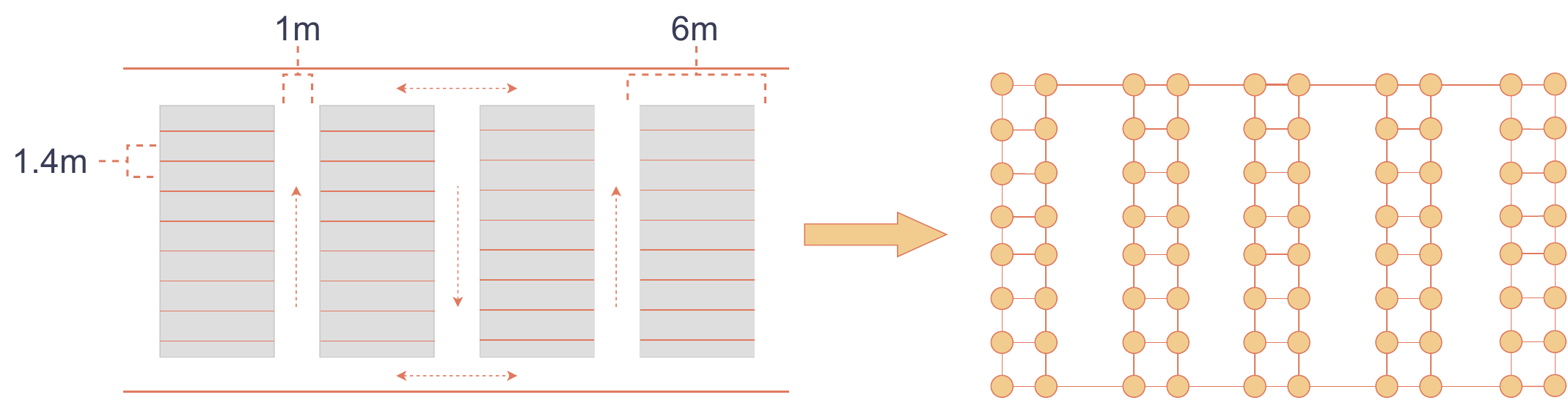}
    \caption{Illustration of the considered warehouse parameters and the associated undirected graph representation of a warehouse. The dotted arrows represent the allowed \acrshort{amr} movement directions, and the grey areas represent storage racks between the aisles. The circles indicate nodes and the connections between the circles are edges.}
    \label{fig: graph representation warehouse}
\end{figure}

The distance between adjacent pick locations within an aisle is 1.4 meters, while the distance to move to the other side of the aisle is 1 meter. The travel distance between two aisles is 6 meters. Figure \ref{fig: graph representation warehouse} clarifies these warehouse parameters.
To enable efficient calculations and to model the warehouse layout, we used a graph structure. Here, the nodes represent locations at which entities can be. The edges represent how entities can move within the warehouse and what the distances between these locations are. We illustrate this graph representation in Figure \ref{fig: graph representation warehouse}. The resulting adjacency and distance matrices can be used to calculate distances and routes within any warehouse layout efficiently. In our use case, we use a directed graph to represent the \gls{amr} travel possibilities, while an undirected graph is used for the pickers, which can move in any direction within the warehouse. Different warehouse structures and travel direction rules can be implemented by switching the graph structure and, therefore, the adjacency and distance matrices.
\par
A pre-generated set of pickruns must be collected by the pickers and \glspl{amr} to fulfill an episode. Thus, simulation episodes start with the pickrun generation and end when all items from all pickruns have been picked. These pickruns contain the list of locations that must be visited, with the number of items that must be picked at each location. The pickrun optimization is a problem on its own. As this is outside our scope, we use a basic approach. Namely, to generate these pickruns, for each episode, we randomly select a set of pick locations. We determine the pickrun lengths using uniform sampling between 15 and 25 locations. We select these lengths based on stakeholder knowledge and the pickrun lengths found in the available data. These pickruns must be collected by the \glspl{amr} via an S-shaped/traversal routing policy. 
Hence, we sort the locations by aisle and then by how early the locations are within their respective aisle, with the aisle entries being on the opposite end of the aisle for consecutive aisles. The picking frequencies at the locations are randomly sampled using the empirical distribution of pick frequencies from the available data. For the pickrun-\gls{amr} assignment, we use a trivial method, with the first pickrun in the queue being assigned to an \gls{amr} that becomes available. 

To start a simulation run, we use a diverse starting method. In diverse starting, all \glspl{amr} are assigned to a pickrun. These pickruns are cut off using a random uniform selection. In this way, the system starts with the \glspl{amr} randomly spread through the warehouse. Similarly, the pickers are randomly allocated to destinations spread throughout the warehouse during instantiation. We do so based on expert knowledge, as initialization procedures to create distributed initial states are common.
To generate a product distribution through the warehouse, we randomly instantiate product locations based on the actual products and product categories in the warehouse data. To do so, for each product category, we gather the distribution of how many items of the category are clustered together. Then, to fill a warehouse with product locations, we randomly sample a product category based on the relative frequencies of the categories. Consequently, we sample how many products must be grouped for this product category based on the empirical distribution. Finally, real-world products of these categories are assigned to these locations. This is done repeatedly until each location contains a specific product with its weight and volume. 
\par
We determine the expected pick time of an order line on a pickrun based on the product characteristics and the number of items that must be picked. To do so, we use an internal method from our industrial partner that was developed using the empirical product and pick time data. 
This method combines the product volume and weight with the number of item pairs and single items that must be picked. Using several empirically tested linear functions that use these two product characteristics and the number of items that must be collected, the expected pick time $t_{\text{pick}}$ can be calculated for each pick. To create the actual pick times, we sampled a value from a Gaussian distribution with $\mu=t_{\text{pick}}$ and $\sigma=0.1\cdot t_{\text{pick}}$.
Since sampling the product characteristics and the number of items that must be picked occurs independently, we verify whether the resulting expected pick times are similar to those calculated from the real order distribution, which contains 100,833 order lines. The histograms of 100,000 sampled picking times through our method and those from the data show a sufficiently similar distribution for our purpose. Besides, the means and standard deviations of the sampled pick times ($\mu=11.3$, $\sigma=10.3$) and of the actual order pick times ($\mu=12.3$, $\sigma=10.8$) are also satisfactory similar.

\subsubsection{Picker Process} \label{sec: pick process simulation}
The picker process describes the picker's logic and how it interacts with the optimizer and \glspl{amr}. The supplementary material contains a schematic overview of this picker process.

In the simulation, we model each transition of one location to another location by a picker or \gls{amr} in the warehouse as an event. This allows us to maintain a detailed overview of the current state of the system with regard to the locations of all pickers and \glspl{amr} at any time. The picker process starts with a picker being allocated to a destination. Once the picker receives its destination, it follows the shortest path to the destination. We set the average picker speed to 1.25 m/s. At the start of each path to a new destination of a picker, we randomly set the speed using a Gaussian distribution with $\mu=1.25$ and $\sigma=0.15$. This mimics the uncertainty in real-world picker speeds. After a timeout of $\text{distance}/ \text{picker speed}$ seconds, the movement event toward the following location takes place. This is repeated until the picker reaches its destination.
\par
When a picker reaches the destination, it checks if any \gls{amr} is waiting there. If no \gls{amr} is waiting at the location, the picker waits until any \gls{amr} arrives there. When an \gls{amr} is waiting at the location or an \gls{amr} arrives, the picking takes place. This picking is represented using a timeout event. The picking time is sampled from a Gaussian distribution. However, in real-world warehouses, picking does not always happen perfectly. Therefore, in consultation with business stakeholders, we include a random picking interruption. Namely, a picking delay is included every once in a while to mimic any uncertainty caused by pickers. This delay can represent pickers having a short break or having to reshuffle items on the \gls{amr}, items being hard to retrieve from the shelve, and so on. We set the frequency of this unexpected delay occurring for each picker using a Poisson random variable with $\lambda=50$, indicating that, on average, a picker has an unexpected delay once per 50 picks.
The distribution fits well when events are independent, which we can assume since a disruption, stock-out, or error at one location generally does not affect those at the next locations. The disruption time is sampled from a Gaussian distribution with $\mu=60$ and $\sigma=7.5$ seconds. 
\par
When a picker has finished picking the items for an \gls{amr}, it checks if any other \gls{amr} is waiting at the location. If so, it picks the items for this \gls{amr}. Then, if no \glspl{amr} are left to be served at the location, the picker must request a new location from the optimizer, and the cycle repeats.

\subsubsection{AMR Process} \label{sec: AMR process}
The \gls{amr} process establishes the behavior of the \glspl{amr} within the system and how they interact with each other and the human pickers. 
An overview of the process logic of the \glspl{amr} in the simulation can found in the supplementary material.
The \gls{amr} process starts with an \gls{amr} being assigned to fulfill a pickrun. Then, like for pickers, an event is used for each transition to the following location on the \gls{amr} path toward the first destination in the pickrun. For \glspl{amr}, these events occur after a timeout of $\text{distance}/ \text{\gls{amr} speed}$ seconds. We set the average \gls{amr} speed to 1.5 m/s. Similar to the picker speed, at each tour toward a new destination, a random speed is selected from a Gaussian distribution with $\mu=1.5$ and $\sigma=0.15$ m/s. Like the pickers, the \gls{amr} continues its movement until it has reached its destination.
\par
However, whereas pickers can walk easily through the warehouse, the \glspl{amr} can encounter congestion. Namely, when another \gls{amr} is standing in the path of the vehicle, it has to do an overtaking maneuver. We model this by adding an overtaking timeout whenever an \gls{amr} has to overtake another \gls{amr}. Since overtaking is a slow procedure for \glspl{amr}, this time penalty is relatively large. We use a random Gaussian sampling method with $\mu=15$ and $\sigma=2.5$ seconds for the overtaking penalty. This penalty indicates the importance of preventing congestion. 
\par
When an \gls{amr} reaches its destination, it waits for the human picker to arrive if it is not yet there. When the human picker arrives, a picking timeout occurs, representing the picker picking the items for the \gls{amr}.
Once an \gls{amr} has been loaded at a pick location, the process is repeated, and the \gls{amr} moves to the next location on its pickrun. This occurs until all locations in the pickrun have been visited. Then, in a real-world warehouse, the \gls{amr} moves to a drop-off location outside the picking area to unload its items. As we do not consider this unloading, we do not include it in our simulation. Instead, the \gls{amr} must return to the base location at the bottom left of the warehouse, where it can start a new pickrun. As the \glspl{amr} at the drop-off location are simply out of the system, not including this process in the simulation does not affect the picking efficiency results. By up- or down-scaling the number of \glspl{amr}, we can still capture the number of \glspl{amr} that are actually in the picking system. 

\subsection{Multi-Objective Markov Decision Process} \label{sec:mdp}
The DRL agent is defined as a \textit{picker allocation optimizer}. We learn a policy that assigns the pickers to pick locations at each step. By doing so, the policy determines the assignment variables $A_{i,k}$ and $U_{i, i'}$.
The picker optimizer receives the system state and allocates the picker to a new destination. Then, the cycle continues until any picker places a new request.
The process is stochastic and multiple pickers usually never place a request at the exact same moment. Therefore, the picker optimizer uses the natural order of incoming requests to allocate pickers one at a time. 
Within this framework, we define the \gls{momdp} below. 

\paragraph{Transition function}  The transition function is formed by the warehouse system. One transition step consists of the picking process between two consecutive allocation requests for the optimizer agent. 
An episode is one warehouse simulation in which a pre-generated set of pickruns is fulfilled. 

\paragraph{State Space}
We use a graph to model the state space, with nodes representing the warehouse locations and edges representing how entities can move between these locations. 
The node features 
are split into two categories: efficiency related and workload fairness related, shown in Tables~\ref{tab:state_space_e} and \ref{tab:state_space}.
For each node, there are 35 node features ($23$ efficiency and $12$ fairness). 

\begin{table}[ht!]
\scriptsize
\renewcommand{\arraystretch}{0.85}
    \centering
    \begin{tabular}{ll}
    \toprule
        \multicolumn{2}{l}{\textbf{Current picker information}} \\ 
        Location & Whether the picker is currently at the node. \\
        \begin{tabular}[l]{@{}l@{}}Picker distance \\\end{tabular}  & \begin{tabular}[l]{@{}l@{}}Provides the distance between picker and the node through warehouse paths.\end{tabular} \\ 
        \midrule
        \multicolumn{2}{l}{\textbf{AMR(s) information}} \\ 
        Location & Whether the AMR is currently at the node. \\
        \begin{tabular}[l]{@{}l@{}} \# of AMRs going \\ \end{tabular}  & \begin{tabular}[l]{@{}l@{}} Number of AMRs currently going towards the node.\end{tabular} \\ 
        \begin{tabular}[l]{@{}l@{}} Destination distance \\ \end{tabular}  & \begin{tabular}[l]{@{}l@{}} Minimum travel distance of AMRs with this node as their destination or -10 if \\none are traveling in towards the node.\end{tabular} \\ 
        \begin{tabular}[l]{@{}l@{}} Expected time until next destination \\ \end{tabular}  & \begin{tabular}[l]{@{}l@{}} Sum of estimated travel time to current destination, pick time at destination and \\time until the next destination. Value of -10 if no AMR goes for the next pickrun,\\ otherwise AMR with minimum travel time is  selected. \end{tabular} \\ 
        \begin{tabular}[l]{@{}l@{}} Expected time until two-step ahead \end{tabular}  & \begin{tabular}[l]{@{}l@{}} Same as expected time until next destination feature but compute the estimates\\ for two-step ahead AMR destination.  \end{tabular} \\ 
        \begin{tabular}[l]{@{}l@{}} \# of AMRs within same aisle \\ \end{tabular}  & \begin{tabular}[l]{@{}l@{}} AMRs going to a destination within the same aisle as the considered node. \\ \end{tabular} \\ 
        \begin{tabular}[l]{@{}l@{}} \# of AMR waiting   \end{tabular}  & \begin{tabular}[l]{@{}l@{}} AMRs currently waiting in the same aisle as  the considered node. \end{tabular} \\ 
        \midrule
        \multicolumn{2}{l}{\textbf{Picker positioning in the system}} \\ 
        \begin{tabular}[l]{@{}l@{}} Location   \end{tabular}  & \begin{tabular}[l]{@{}l@{}} Indicate if any picker other than the picker being assigned is at this node. \end{tabular} \\ 
        \begin{tabular}[l]{@{}l@{}} Minimum travel distance  \end{tabular}  & \begin{tabular}[l]{@{}l@{}} Minimum distance to this node among all pickers having this node as destination.\\ If none, the value is -10. \end{tabular} \\ 
        \begin{tabular}[l]{@{}l@{}} \# of pickers  \end{tabular}  & \begin{tabular}[l]{@{}l@{}} Number of pickers going to a destination within the same aisle as the considered\\ node. \end{tabular} \\ 
        \begin{tabular}[l]{@{}l@{}} Distance of other pickers  \end{tabular}  & \begin{tabular}[l]{@{}l@{}} Minimum distance of any other picker to its current destination plus the
distance \\from its current destination to the considered node. \end{tabular} \\ 
        \begin{tabular}[l]{@{}l@{}} Expected time of  other pickers  \end{tabular}  & \begin{tabular}[l]{@{}l@{}} Similar to the above, but considering the expected time, including expected\\ picking time at the current destination.  \end{tabular} \\ 
         \midrule
        \multicolumn{2}{l}{\textbf{Node region information}} \\
        \begin{tabular}[l]{@{}l@{}} Aisle distance from origin  \end{tabular}  & \begin{tabular}[l]{@{}l@{}} How far the aisle of this node is from the origin, scaled by the warehouse size. \end{tabular} \\ 
        \begin{tabular}[l]{@{}l@{}} Node depth within aisle  \end{tabular}  & \begin{tabular}[l]{@{}l@{}} How far toward the beginning or end of the aisle a node is located, scaled by the\\ aisle length. \end{tabular} \\ 
        \midrule
        \multicolumn{2}{l}{\textbf{Node neighborhood features}} \\
        \begin{tabular}[l]{@{}l@{}} Closest next destination distances \end{tabular}  & \begin{tabular}[l]{@{}l@{}} Closest and \nth{2} closest distance to the next destinations of the \glspl{amr} going to\\ this node. 0 if no \glspl{amr} or last node in the pickrun. \end{tabular} \\ 
        \begin{tabular}[l]{@{}l@{}} Closest distances to two-step ahead.  \end{tabular}  & \begin{tabular}[l]{@{}l@{}} Same as above but for the closest two-step ahead destination. \end{tabular} \\ 
        \begin{tabular}[l]{@{}l@{}} Closest distance to pickers  \end{tabular}  & \begin{tabular}[l]{@{}l@{}} Minimum distances from this node to the other nodes that are currently the \\destination of any of the pickers.   \end{tabular} \\ 
        \begin{tabular}[l]{@{}l@{}}  Distances to closest unserved AMRs \end{tabular}  & \begin{tabular}[l]{@{}l@{}} Distances to the closest and \nth{2} closest other nodes that are the destination of an\\ AMR and where no picker is
        already going. \end{tabular}\\
        \bottomrule
    \end{tabular}
    \caption{List of state space features related to efficiency.}
    \label{tab:state_space_e}
\end{table}

\begin{table}[ht]
\scriptsize
\centering
\renewcommand{\arraystretch}{0.9}
    \begin{tabular}{ll}
    \toprule
        \multicolumn{2}{l}{\textbf{Node specific workload information }} \\
        \begin{tabular}[l]{@{}l@{}} Current picker workload  \end{tabular}  & \begin{tabular}[l]{@{}l@{}} Total mass in kilograms that the picker at this node has picked subtracted by the mean\\ workload of all pickers.  \end{tabular} \\ 
        \begin{tabular}[l]{@{}l@{}} Next picker workload   \end{tabular}  & \begin{tabular}[l]{@{}l@{}} Same as above when the picker destination is the considered node.  \end{tabular} \\ 
        \begin{tabular}[l]{@{}l@{}} Item weight  \end{tabular}  & \begin{tabular}[l]{@{}l@{}} Mass in kilograms of a single item stored at the node.  \end{tabular} \\ 
        \begin{tabular}[l]{@{}l@{}}  Waiting AMR workload \end{tabular}  & \begin{tabular}[l]{@{}l@{}} Mass of the items that must be loaded on the waiting AMRs at this location. \end{tabular} \\ 
        \begin{tabular}[l]{@{}l@{}} Destination AMRs workload  \end{tabular}  & \begin{tabular}[l]{@{}l@{}} Mass of the items that must be loaded on the AMRs that are going to this location but\\ are not yet there.  \end{tabular} \\ 
        \begin{tabular}[l]{@{}l@{}} Closest picker workloads \end{tabular}  & \begin{tabular}[l]{@{}l@{}} Total masses carried by the two closest pickers to this node in terms of expected arrival\\ time, subtracted by the mean picker workload.  \end{tabular} \\ 
        \midrule
        \multicolumn{2}{l}{\textbf{Distributional workload information }} \\
        \begin{tabular}[l]{@{}l@{}} Picker total workload   \end{tabular}  & \begin{tabular}[l]{@{}l@{}} Workload in kilograms of the controlled picker 
        subtracted by the mean picker workload.  \end{tabular} \\ 
        \begin{tabular}[l]{@{}l@{}} Other picker workloads \end{tabular}  & \begin{tabular}[l]{@{}l@{}} Minimum, \nth{25} and \nth{75} percentile, maxixmum workload of all pickers, subtracted by the\\ mean picker workload.  \end{tabular} \\ \bottomrule
    \end{tabular}
    \caption{List of state space features related to workload fairness.  }
    \label{tab:state_space}
\end{table}

The efficiency related features consist of information of the current picker, \glspl{amr}, other pickers, node location, and node neighborhood.
The current picker information describes the positioning of the nodes in relation to the controlled picker for whom an allocation decision must be made. 
The \gls{amr} information describes the positioning and next destinations of the \glspl{amr} with respect to the nodes. This helps to identify promising picking locations based on the \gls{amr} distribution in the warehouse.
For similar reasons, we also included 5 node features describing other picker information in the state space. This allows the \gls{drl} agent to consider other pickers' actions and locations in the decision-making process, which can prevent unnecessary picker overlap and aid synergies.
The node region information describes the regions in which the nodes are located within the warehouse. These features may help policies consider the routing of \glspl{amr}. Namely, if two nodes are within the same aisle, it may be beneficial to first pick the one at the aisle entry since the \gls{amr} will continue its route toward the aisle end. Similarly, if two nodes are in consecutive aisles, picking the node in the aisle closer to the start could be beneficial. 
Lastly, we selected the node neighborhood features to capture the picking process occurring around the nodes. These may help capture 
high- or low-density pick areas. 

The workload fairness features are split into node-specific features describing the workload characteristics at the nodes and ``distributional'' features describing the current distribution of picker workloads. Although the distributional features are not node-specific, we included them as node features to facilitate node-wise computations. Thus, these features contain the same value for each node. These features allow for consideration of workloads at specific picking locations while also considering the current picker workload in comparison to the overall workload distribution.

\paragraph{Action Space}
We use a discrete action space that consists of the nodes in the graph. Namely, a policy should assign a picker that places an allocation request to a single node, representing the new picker destination. We use a truncated action space that, at any timestep, consists of all locations that are the current or the next destination in any \gls{amr} pickrun and where no other human picker is already going. The maximum size of this variable action space is achieved when all the \gls{amr} destinations and next destinations are unique locations. Then, the action space has a size of $2 \times \text{nr. of \glspl{amr}} - (\text{nr. of pickers} - 1)$.

\paragraph{Reward Function}
We use one reward signal for efficiency and one for workload fairness. For efficiency, we use a penalty on the passed time. Specifically, at each transition step, the penalty is the elapsed time in seconds between the current step and the previous step. Formally, the reward is as follows, with $T_t$ indicating the system time at step $t$: $R_t^{\text{efficiency}} = T_{t-1} - T_t$.

For fairness, the reward at each step is based on the increase or decrease of the standard deviation of the total carried product masses between the previous and current steps. So, at step $t$, the fairness reward is as follows, with  the standard deviation $\sigma$, $W_{k,t}$ indicating the total lifted mass by picker $k$ until step $t$, and $|\mathcal{K}|$ the number of pickers: 
$R_t^{\text{fairness}} = \sigma(W_{1,t-1}, \dots, W_{|\mathcal{K}|,t-1}) -\sigma(W_{1,t}, \dots, W_{|\mathcal{K}|,t})$. 
The output vector of the reward function in the \gls{momdp} at each step $t$ is: $\mathbf{R}_t= (R_t^{\text{efficiency}}, R_t^{\text{fairness}})$.

\subsection{Aisle-Embedding Multi-Objective Aware Network} \label{sec:aemo_net}
\begin{figure}[th]
    \centering    \includegraphics[width=0.7\textwidth]{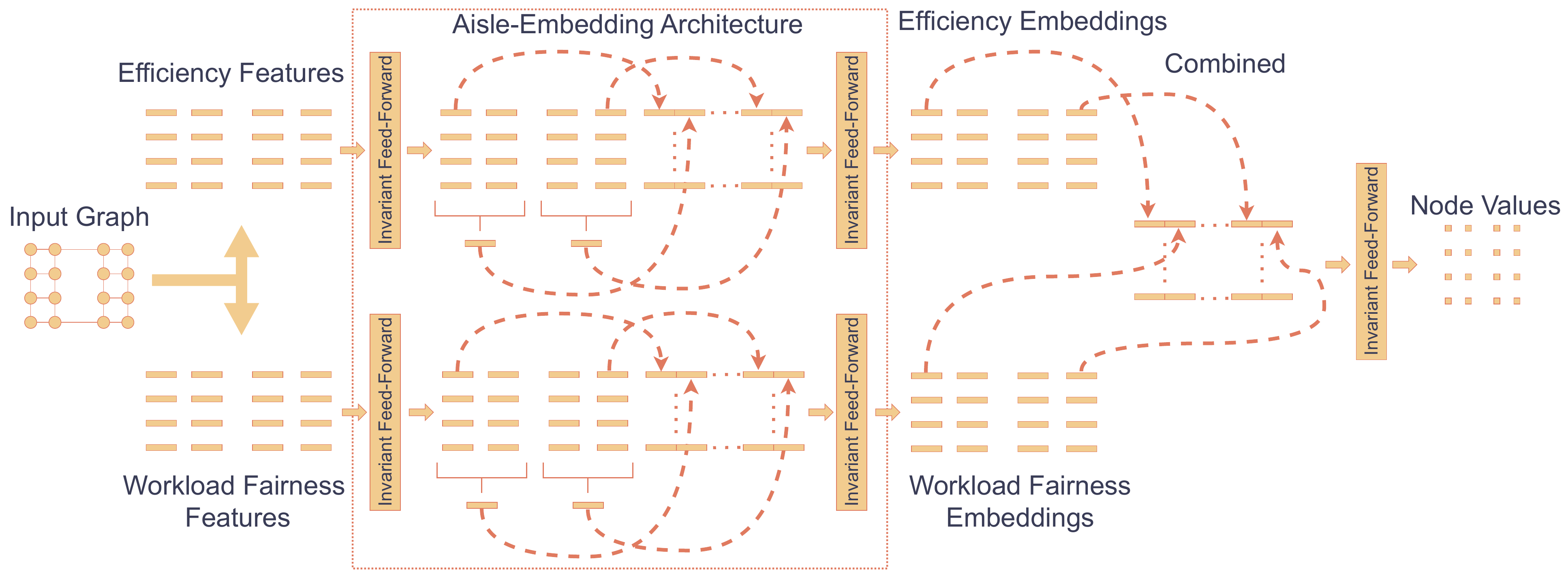}
    \caption{Illustration of the AEMO-Net architecture.}
    \label{fig: feature separation actor}
\end{figure}

We propose an Aisle-Embedding Multi-Objective Aware Network (AEMO-Net), a graph-based architecture tailored to capture neighborhoods within deep warehouse aisles comprising of often 30-40 nodes.  The standard message-passing method of graph neural networks falls short in larger node settings~\citep{balcilar2021breaking} resulting in multiple message-passing steps and deep networks which are difficult and slow to learn. 
Figure~\ref{fig: feature separation actor} outlines the proposed architecture. The aisle-embedding structure combines the idea of permutation invariant aggregation from graph networks with our warehouse domain knowledge. Specifically, aisles form natural regions of related nodes within a warehouse. By aggregating the embeddings of the nodes within an aisle, we create an aisle-embedding that captures the regional information. Then, we combine the node-embedding with the aisle-embedding to calculate the final node values used to output the action probabilities. Formally, the aisle-embedding of an aisle $A$ is calculated as follows, with $\boldsymbol{h}_v^l$ indicating the node embeddings at layer $l$, and $\mathcal{V}_A$ the set of nodes within an aisle $A$: $\boldsymbol{h}_{A}^l = \Psi\left(\{\boldsymbol{h}_v^l | v\in \mathcal{V}_A  \}\right).$
We use the mean as the permutation invariant function $\Psi$.
 
To facilitate multi-objective learning, we use an architecture that separates the two feature categories and treats them independently before their high-level embeddings are combined. This enables learning embeddings related to both feature categories without noise while the shared final layers capture the interactions between the fairness and efficiency objectives. Combining the aisle-embedding structure with feature separation, the AEMO-Net is formulated as: 
$AEMO(v) = \gamma_{\text{actor}} \left(\left[Emb_{\text{fair}}(v), Emb_{\text{effic}}(v) \right] \right)$. 
Here,
$Emb_\text{cat}$ represents the aisle-embedding network for a feature category and $\mathbf{x}_{v}^\text{cat}$ the feature vector of a category for node $v$: 
$
    Emb_\text{cat}(v) = \phi_{\text{actor}}^\text{cat}([\psi_{\text{actor}}^\text{cat}(\mathbf{x}_{v}^ \text{cat}),\allowbreak \text{AVG}(\{\psi_{\text{actor}}^\text{cat}(\mathbf{x}_{u}^\text{cat}) | u \in \mathcal{V}_{\text{aisle}({v})}\})])$. 
For the critic network, we do not use the aisle-embedding architecture because preliminary tests showed that it is not required to approximate the value function well. 
Instead, we use feature separation with invariant feed-forward layers: 
$$
Critic(G) = \gamma_{\text{crit}} \left(\sum\left( \left\{ \phi_\text{crit}\left(\left[\psi_{\text{crit}}^\text{effic}(\mathbf{x}_v^\text{effic}), \psi_{\text{crit}}^\text{fair}(\mathbf{x}_v^\text{fair})  \right]\right)| v \in \mathcal{V}_G \right\}   \right)\right).
$$

\subsection{Multi-objective Learning Algorithm} \label{sec:learning_algorithm}
We extend the multi-objective RL algorithm in \cite{Xu2020} to handle discrete action spaces and graph state spaces. Algorithm \ref{alg: pgmorl} shows the pseudo-code.

\begin{algorithm}[ht]
   \caption{Multi-Objective Learning Algorithm.} \label{alg: pgmorl}
\setstretch{1.0}
\begin{algorithmic}[1]
\small
\Require Nr. parallel tasks $n$, Nr. warm-up iterations $m_w$, Nr. task iterations $m_t$, Nr. generations $M$.
\State Initialize population $\mathcal{P}$, Pareto archive $EP$, and \acrshort{rl} history $\mathcal{R}$.
\LineComment{Warm-up Phase}
\State Generate initial task set $\mathcal{T} = \{\pi_j, \boldsymbol{\omega}_j\}_{j=1}^n$ using random policies $\pi_j$ and evenly distributed weight vectors $\boldsymbol{\omega}_j$.
\For{task $(\pi_j, \boldsymbol{\omega}_j) \in \mathcal{T}$} \Comment{Run in parallel.}
\State Run \acrshort{ppo} for $m_w$ iterations.
\State Collect result policy $\pi_j'$ and intermediate policies in $\mathcal{P}'$
\State Store eval. rewards of old, new, and intermediate policies with weights $\boldsymbol{\omega}_j$ in $\mathcal{R}$
\EndFor
\State Update $\mathcal{P}$ and $EP$ with $\mathcal{P}'$.
\LineComment{Evolutionary Phase}
\For{$generation \gets 1,2,\dots, M$}
\State Fit improvement prediction models for each policy in $\mathcal{P}$ using data in $\mathcal{R}$
\State Select new task set $\mathcal{T} = \{\pi_j, \boldsymbol{\omega}_j\}_{j=1}^n$ based on improvement predictions.
\For{task $(\pi_j, \boldsymbol{\omega}_j) \in \mathcal{T}$} \Comment{Run in parallel.}
\State Run \acrshort{ppo} for $m_w$ iterations.
\State Collect result policy $\pi_j'$ in $\mathcal{P}'$
\State Store eval. rewards of old, new, and intermediate policies with weights $\boldsymbol{\omega}_j$ in $\mathcal{R}$
\EndFor
\State Update $\mathcal{P}$ and $EP$ with $\mathcal{P}'$.
\EndFor
\end{algorithmic}
\end{algorithm}
The core concept is to learn \gls{drl} policies using \gls{ppo} \citep{Schulman2017} training with a weighted-sum reward function $R_t = \boldsymbol{\omega}^T \mathbf{R}_t$, with $\boldsymbol{\omega}$ a weight vector and $\mathbf{R}_t$ the reward vector at time $t$. The algorithm steers learning toward the weight vectors expected to stimulate policies that improve the current non-dominated set of solutions. 
To do so, the algorithm starts with a warm-up phase, where 
$n$ tasks are initialized. A task $j$ consists of a policy $\pi_j$ and a weight vector $\boldsymbol{\omega}_j$. The initial tasks consist of randomly initialized policy networks and evenly distributed weight vectors between 0 and 1. 
These initial tasks are trained using \gls{ppo} for $m_w$ warm-up iterations. The trained policies, intermediate policies, and their evaluation rewards are stored in a population $\mathcal{P}$ of both non-dominated and dominated policies. Based on the evaluation rewards, the intermediate Pareto archive is also updated to contain the non-dominated solutions. Thus, the warm-up phase outputs several baseline policies for different objective preferences.
\par
Then, in the evolutionary phase,
at each generation, for each policy in the population $\mathcal{P}$, a prediction model is made to predict the rewards that can be achieved if the policy is trained using a specific weight vector. This four-parameter hyperbolic model for each policy and objective function is trained based on data samples stored in history $R$ that are in the neighborhood of the policy.
Using this prediction model, tasks (i.e., policies combined with a weight vector) are selected such that the predicted new non-dominated set improves the most, based on the hypervolume and sparsity. 
Consequently, \gls{ppo} training is done for $m_w$ iterations, and the results are stored. Then, the evolutionary cycle repeats. The final output is a set of non-dominated policies. 

For the internal \gls{ppo} training, we use the actor-critic variant with the clipped loss function and entropy term. Algorithm \ref{alg:PPO algorithm} outlines the \gls{ppo} algorithm. \gls{ppo} is a so-called policy-based algorithm used to train a policy neural network $\pi$ to output action probabilities. To do so, the algorithm alternates between collecting samples and updating the policy using the empirical estimates from these samples. The loss function $\mathcal{L}^{CLIP}$ used to update the network is as follows.

\begin{algorithm}[ht]
\caption{PPO learning algorithm.} \label{alg:PPO algorithm}
\begin{algorithmic}[1]
\small
\Require Number of iterations $N$, initial actor parameters $\theta_0$, initial critic parameters $\phi_0$.
\State $i \gets 0$
\While{$i<N$}
    \State Collect trajectories by running policy $\pi_i=\pi(\theta_i)$ in parallel environments.
    \State Compute advantage estimates $\hat{A}_t$ using critic network $V_i=V(\phi_i)$.
    \State Update policy $\theta_{k}$ to $\theta_{k+1}$ via gradient descent on PPO loss $\mathcal{L}(\theta_{k})$.
    \State Update critic $\phi_k$ to $\phi_{k+1}$ via gradient descent on mean-squared error loss.
\EndWhile
\end{algorithmic}
\end{algorithm}

\[\mathcal{L}^{CLIP}(\theta)= \hat{\mathbb{E}}_t \left[\min\left(\frac{\pi_\theta(a_t|s_t)}{\pi_{\theta_\text{old}}(a_t|s_t)} \hat{A}_t, \text{clip}\left(\frac{\pi_\theta(a_t|s_t)}{\pi_{\theta_\text{old}}(a_t|s_t)}, 1 - \epsilon, 1 + \epsilon\right) \hat{A}_t \right)\right]\]
Here, $\hat{\mathbb{E}}_t$ indicates the empirical expectation based on the collected samples, $\frac{\pi_\theta(a_t|s_t)}{\pi_{\theta_\text{old}}(a_t|s_t)}$ describes the ratio between the probabilities of the old and new policy of selecting action $a_t$ in state $s_t$ at time $t$, and $\hat{A}_t$ is an estimator of the advantage function at time $t$, indicating how good the action taken at time $t$ was. Thus, the loss tries to maximize the probability of taking good actions and minimize the probability of taking bad actions. To ensure the policy does not change too drastically, the ratio is clipped using the hyperparameter $\epsilon$, limiting the loss. The advantage function $\hat{A}_t$ is estimated by the critic network that is updated during the learning process. To handle the exploration-exploitation trade-off within the \gls{ppo} algorithm, the loss function includes an entropy term. This term measures the spread of the probabilities. This is incorporated as follows.
\[\mathcal{L}(\theta) = \hat{\mathbb{E}}_t \left[ \mathcal{L}^{CLIP}(\theta) + c_{ent} \cdot S[\pi_\theta](s_t)\right]\]
Here, $c_{ent}$ is the entropy coefficient, which determines the weight of the entropy within the loss function, and $S[\pi_\theta](s_t)$ represents the entropy measure. By choosing a small value $c_{ent}$, the algorithm focuses more on exploitation instead of exploration when the clipping loss has been reduced.

\section{Experiments} 
In this section, we first introduce the baselines that we use to compare our method and define the implementation details of our method. Then, we perform initial single-objective experiments, showcasing the quality on the efficiency objective, followed by elaborate multi-objective experiments.
We define various scenarios with different warehouse sizes and picker/AMR ratios. 
These scenarios are used to evaluate learning performance on problems of different
scales and situations, but also how the learned policies transfer directly to different environments. Table \ref{tab: warehouse types} gives an
overview of the basic warehouse scenarios. Note that the $XL$ type resembles the size of a large supermarket warehouse in practice. 

\begin{table}[h]
\renewcommand{\arraystretch}{0.95}
\small
\setlength{\tabcolsep}{5.8pt}
\centering
\begin{tabular}{lcccccc}
\toprule
Type & Aisles & Depth & \# Loc. & Pickers & \acrshortpl{amr} & Picks \\ \midrule
$S$ & 10 & 10 & 200 & 10 & 25& 5000 \\
$M$ & 15 & 15 & 450 & 20 & 50 &7500\\
$L$ & 25 & 25 & 1250 & 30 & 90 &7500\\
$XL$ & 35 & 40 & 2800 & 60 & 180 &15000\\
\bottomrule
\end{tabular}
\caption{Overview of warehouse types in the experiments.}\label{tab: warehouse types}\end{table}
\subsection{Baseline and Benchmark Methods}
 We first implement the previously defined \gls{milp} model without the workload fairness considerations (i.e., without Equation 13 and fairness objective). This baseline is used to compare our method on small, deterministic, single-objective instances.
For larger, stochastic instances with fairness considerations, the \gls{milp} method cannot be utilized. Hence, we use two benchmark methods. First, the greedy
baseline that
always assigns a picker to the nearest
available location where an AMR is going and no other picker is already going. 
The second benchmark (referred to as VI benchmark) reflects our industrial partner's current method. Under this rule-based approach, a picker scans 10 locations ahead or behind in an aisle to find awaiting AMRs. If any are found, the picker moves to the nearest one. Among multiple AMRs, the priority goes to the one encountered first. If no AMRs are found in the scanned area, the picker takes a step in the allowed AMR travel direction, and the process repeats. When the picker reaches the aisle's end, they are reassigned to a new aisle by selecting the aisle with the lowest cost:
$\text{Aisle Cost} = \text{Nr. of aisles difference} - \text{Nr. of waiting AMRs}$. 
Consequently, the picker moves to this aisle, where the process is repeated.
\subsection{Network Architectures}
In the actor-network, to create the efficiency and fairness embeddings, we use two fully-connected layers with 64 neurons and the Leaky ReLU activation function. These are followed by a fully-connected layer with 16 neurons. The Leaky ReLU  $\alpha=0.01$ for all models. The output is used to create the aisle-embeddings, and the node- and aisle-embeddings are stacked to get a 32-dimensional node representation. To create the node efficiency and fairness embeddings, we use two fully-connected layers with Leaky ReLU activation and 64 and 16 neurons, respectively. The 16-dimensional embeddings are stacked to create 32-dimensional combined node embeddings. A final fully-connected layer with 16 channels and Leaky ReLU is followed by a single neuron layer. These final node values are masked by setting their values to negative infinity, and the softmax function is applied to get the action probabilities.
\par
In the critic network, 
we use the same principle of applying the same architecture to both the efficiency and workload fairness features. Thus,
we use the same three fully-connected layers per feature category. The resulting embeddings with 16 layers are stacked to form a 32-dimensional embedding. These 32-dimensional embeddings are passed through a 16-neuron fully-connected Leaky ReLU layer and summed to get the aisle embedding. Then, one final linear layer of 2 neurons outputs the two value estimates.

\subsubsection{Pure Efficiency and Fairness Networks}
We use an actor network with aisle-embedding structure and the critic network with an invariant feed-forward encoder. In the actor network architecture, to generate the node-embeddings, we use two fully-connected layers with 64 neurons and the Leaky ReLU activation function, followed by a fully-connected layer with 16 neurons. The output is used to create the aisle-embeddings, and the node- and aisle-embeddings are stacked to get node representation vectors of length 32. To create the final node values from the vectors, we use two fully-connected layers with Leaky ReLU activation and 64 and 16 neurons, respectively, followed by a single-neuron fully-connected layer. Then, invalid nodes are masked, and the softmax function is used to get the action probabilities.
\par
In our critic networks, we use three fully-connected layers with the Leaky ReLU activation function to create the node-embeddings. For the first two layers, we use 64 neurons, while the third layer has 16 neurons. Then, after aggregating the node-embeddings to form the graph-embedding, we use one fully-connected layer with one neuron to get the value estimate.

\subsection{Learning Algorithm}
For the \gls{ppo}, we use 64 parallel environments with 400 collected experience tuples per environment per \gls{ppo} iteration. For the loss function, we set the clipping parameter $\epsilon$ to $0.2$ and entropy coefficient $c_{ent}$ to $0.01$ after preliminary tests. We use the Adam optimizer \citep{Kingma2014} with a learning rate of $5\times 10^{-4}$. Per \gls{ppo} iteration, we perform three epochs with a batch size of $128$. We set the discount factor $\gamma$ to $0.995$. During training, we sample the actions of the policies based on the output probabilities, to facilitate exploration.
We always pick the actions with the highest action probability for evaluation.

For the multi-objective learning algorithm, we use 6 parallel tasks. For warehouse type $S$, we set the number of warm-up iterations $m_w$ to 80 and the number of task iterations between evolutionary steps $m_t$ to 12. For warehouse types $M$ and $L$, we set $m_w$ and $m_t$ to $128$ and $16$, respectively. For warehouse $S$, we collect $7$ million steps per task before termination, while for warehouse types $M$ and $L$, we use $7.5$ million steps per task before termination. We perform 20 evaluation episodes once every $6$ and $8$ \gls{ppo} for type $S$, and types $M$ and $L$, respectively. 
Lastly, we normalize both reward functions to similar scales. Although not strictly necessary, this aids in finding better weight vectors oppositely to when rewards are of different magnitudes. For all other algorithm settings, we use the values defined by \citet{Xu2020}.
\par
To train the pure efficiency and fairness policies, we use PPO training using the same parameters. We train for 150 epochs for warehouse type $S$, 200 epochs for types $M$ and $L$, and 400 epochs for type $XL$, which shows convergence. To train the pure efficiency policies, we only include the efficiency related features and reward, while for pure fairness policies, we only include the workload fairness related features and reward. We train all policies on a machine with a 32-core Intel Xeon Platinum 8360Y processor and an NVIDIA A100 GPU.

\subsection{Single-Objective Results}
To assess the quality of our proposed method, we first evaluate the single-objective efficiency performance. To do so, we train policies for all previously mentioned warehouse sizes. Then, we we run 100 evaluation episodes per policy on the same warehouse type as they were trained on. Each episode has a unique set of pickruns and allocation of products through the warehouse. We compare the results with the benchmark methods.
\par
Table \ref{tab: performance evaluation} shows the results. We find that the \gls{drl} policies outperform the greedy and VI benchmark policies by a clear margin for all warehouse sizes. For the smallest warehouse size, the performance improvement over the VI benchmark is 14.9\% percent, while for the larger warehouse sizes \gls{drl} achieves over 30\% faster completion times, with improvements of 31.7\% and 33.6\% for warehouses $L$ and $XL$, respectively. The greedy baseline performs slightly worse than the VI Benchmark, although the differences are just a few percent. These findings demonstrate that the \gls{drl} policies perform well as picker optimizer agents in collaborative order picking warehouses and that they can achieve good efficiency in realistically-sized warehouse instances with randomness, congestion, and unexpected interruptions.

\begin{table}[h]
\centering
\small
\begin{tabular}{lccccc}
\toprule
 & \multicolumn{2}{c}{\acrshort{drl}} & \multicolumn{2}{c}{Greedy} & VI Benchmark \\ \cmidrule{2-6}
Warehouse & Picking Time & $\%$ & Picking Time & $\%$ & Picking Time \\ \midrule
$S$ & $\mathbf{8586 \pm 62}$ & $\mathbf{14.9}$ & $10619 \pm 59$ & $-5.3$ & $10087 \pm 58$ \\
$M$ & $\mathbf{8425\pm 46}$ & $\mathbf{21.0}$ & $11023 \pm 58$ & $-3.3$ & $10669 \pm 41$ \\
$L$ & $\mathbf{6540 \pm 37}$ & $\mathbf{31.7}$ & $9823 \pm 33$ & $-2.7$ & $9569 \pm 61$ \\ 
$XL$ & $\mathbf{9010 \pm 21}$ & $ \mathbf{33.6} $ & $13972 \pm 44$ & $ -3.0 $ & $13570 \pm 72$ \\\bottomrule
\end{tabular}
\caption{Performance evaluation on picking efficiency. The values indicate the average picking time in seconds over 100 evaluation episodes, with $\pm$ indicating the width of the 95\%-confidence intervals. The $\%$ indicates the percentage improvement over the VI Benchmark, with a positive percentage indicating an improvement and, thus, lower picking times. The bold markings indicate the best performance values per warehouse size.}\label{tab: performance evaluation}
\end{table}
\subsubsection{Deterministic Instance Evaluation}  \label{sec:deterministic_evaluation}
In addition to the previous results, we perform additional experiments to understand how close we can get to optimal results. To do so, we test several warehouse instances with fully deterministic settings. We use fixed picking times of 7.5 seconds, fixed picker and \gls{amr} speeds of 1.25 m/s and 1.5 m/s, respectively, no overtaking penalties, and no random disruptions. The warehouses have 7 aisles with a depth of 7 (98 picking locations) and we include 4 pickers and 7 \glspl{amr}. The instances we use all contain one pickrun per \gls{amr}. For each instance, we sample random pickruns of lengths between 9 and 14 items. We test two different instance types. First, we test instances with diverse starting positions in which we cut off the sampled pickruns using random uniform selection to ensure that \glspl{amr} are spread through the warehouse. Second, we test instances without diverse starting positions. In these instances, all \glspl{amr} start a full pickrun, meaning they are initialized closer to each other at the beginning of the warehouse.
\par
We train one \gls{drl} agent for the diverse starting scenarios and one for the non-diverse starting scenarios. In training, we use random warehouse instantiations with the same overall warehouse parameters. Thus, the \gls{drl} policies were not explicitly trained for the specific testing instances. We evaluate the \gls{drl} policies on each evaluation instance. In addition, we evaluate the greedy and VI Benchmark methods on these instances.
\par
We implement and solve the equivalent \gls{milp} instances using the Gurobi solver \citep{gurobi}. We use their indicator constraints option to solve the constraints with big-$M$ notation as efficiently as possible. For each instance, we run the Gurobi solver for 20 hours on a computer with an AMD Rome 7H12 CPU instance with 64 CPU cores.
\par
Table \ref{tab: MILP comparison} shows the results. The first thing that stands out is that the solver could not prove optimality within 20 hours, as indicated by the \gls{milp} gap. This indicates the complexity of the problem, even in these minimalistic, deterministic instances.
\begin{table}[h]
\small
\centering
\begin{subtable}{\linewidth}
\centering
\begin{tabular}{lccccc}
\toprule
Instance & \acrshort{drl} & Greedy & VI Benchmark & \acrshort{milp} & \acrshort{milp} gap (\%)\\ \midrule
1 & 154 & 154 & 355 & \textbf{149} & 17.8\\
2 & \textbf{187} & 190 & 397 & \textbf{187} & 6.0\\
3 & 155 & 167 & 299 & \textbf{149} & 12.2\\
4 & \textbf{206} & 248 & 269 & 212 & 17.5\\
5 & 227 & 236 & 277 & \textbf{206} & 15.9\\ \bottomrule
\end{tabular}
\caption{Instances with diverse starting.}
\end{subtable}
\begin{subtable}{\linewidth}
\centering
\begin{tabular}{lccccc}
\toprule
Instance & \acrshort{drl} & Greedy & VI Benchmark & \acrshort{milp} & \acrshort{milp} gap (\%) \\ \midrule
1 & \textbf{244} & 262 & 355 & \textbf{244} & 28.2\\
2 & \textbf{249} & 253 & 297 & 271 & 28.1\\
3 & \textbf{265} & 272 & 299 & 267 & 29.3\\
4 & \textbf{240} & 257 & 269 & 245 & 22.8\\
5 & \textbf{251} & 255 & 277 & 260 & 30.9\\ \bottomrule
\end{tabular}
\caption{Instances without diverse starting.}
\end{subtable}
\caption{Performance evaluation for multiple small, deterministic warehouse instances. The values indicate the total picking time in seconds for the specific problem instance. The \acrshort{milp} gap indicates the percentage gap between the lower bound estimate of the solver and the best found solution. The bold markings indicate the best performance values per problem instance.}\label{tab: MILP comparison}
\end{table}

\par
We also find that the \gls{drl} solutions are very close to the best \gls{milp} solution in all cases. \gls{drl} even achieves better results for 5 instances. The biggest deviation in total picking time from the best \gls{milp} solution is just 21 seconds (227 vs. 206), indicating that \gls{drl} policies can consistently achieve good results. In addition, the \gls{drl} agents outperform the greedy and VI benchmark methods for each instance. Compared to the greedy baseline, the improvement is generally not large. However, with such small instances, no congestion, and the results being so close to the \gls{milp} results, we cannot expect a large deviation from the greedy method. That is, the greedy method optimizes in the short run without much consideration of other pickers, leading to fast initial picks for the pickers. In such short episodes, the long-term consequences cannot be affected too much as episodes end relatively quickly. In addition, the greedy method experiences the converse effects of congestion less due to the lack of overtaking penalties. The VI benchmark results are worse than greedy and \gls{drl}. This makes sense as this method was developed to spread the pickers more evenly through the warehouse, while this may be less beneficial in short episodes without congestion effects. All in all, the deterministic instance results show that we can achieve good, near-optimal solutions using \gls{drl} that match the performance of the best solutions found by a solver with complete information of the problem instances.

\subsection{Multi-Objective Results}
We train policies for warehouse types $S$, $M$, and $L$. This results in a set of non-dominated policies for each type. We gather these policies and run 100 evaluation episodes per policy on the same warehouse type as they were trained on. The obtained policies are compared in terms of total picking time and the standard deviation of the workloads. We further compare them with the two baselines, i.e., pure efficiency, and pure fairness.

\begin{figure}[ht!]
\begin{subfigure}{0.32\textwidth}
    \centering
    \includegraphics[width=\linewidth]{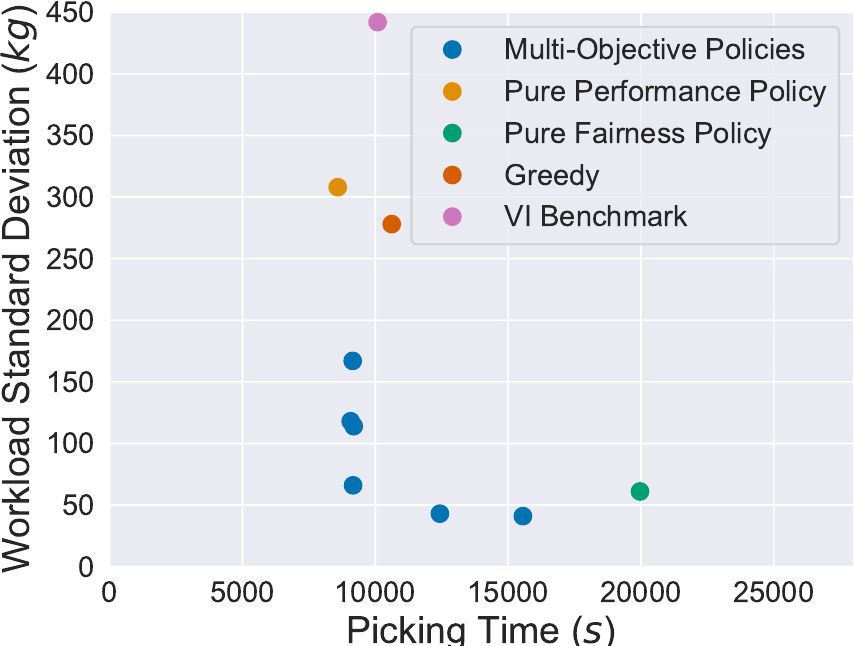}
    \caption{Type $S$.}
    \label{fig:morl_results_type_S}
\end{subfigure}\hfill
\begin{subfigure}{0.33\textwidth}
    \centering
\includegraphics[width=\linewidth]{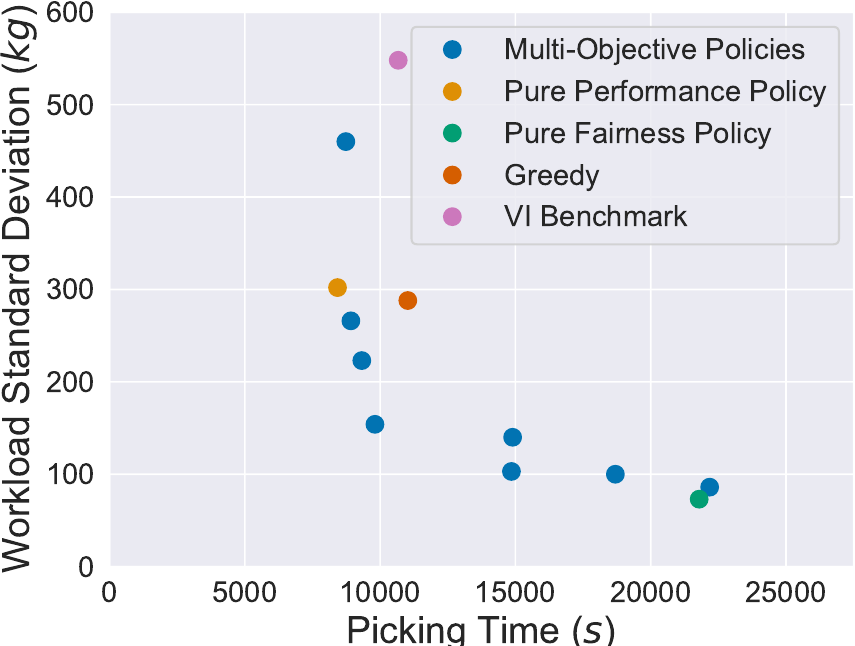}
    \caption{Type $M$.}
\end{subfigure}\hfill
\begin{subfigure}{0.33\textwidth}
    \centering
\includegraphics[width=\linewidth]{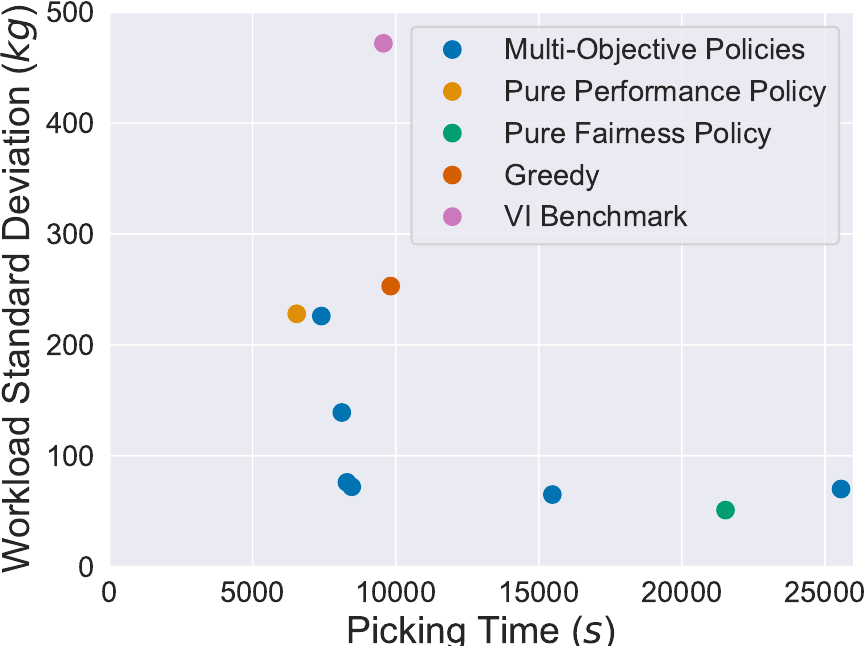}
    \caption{Type $L$.}
\end{subfigure}
\caption{Performance of the non-dominated sets of policies learned on warehouse different warehouse types.}
\label{fig: morl_results}
\end{figure}

Table \ref{tab: MORL performance evaluation} and Figure \ref{fig: morl_results} show the performance of non-dominated policies on different warehouse types. 
There are 6 non-dominated policies for sizes $S$ and $L$,
whereas, 8 for size $M$. 
In Figure \ref{fig:morl_results_type_S}, the non-dominated set of multi-objective policies forms a clear front toward the bottom left. The policies show a trade-off with a relatively sharp ``angle.'' This shows that we can decrease the workload standard deviation a lot before we sacrifice much pick efficiency or decrease the picking time by a lot before the workload fairness deteriorates. A policy that stands out is policy S3, which is represented by the dot in the bottom left of the front. This policy achieves both good completion times and good workload fairness. Namely, the average time to complete an episode is 9164 seconds, and the workload standard deviation is 66 kilograms, compared to 8586 seconds and 308 kilograms of the pure performance policy. Thus, by sacrificing just 6.7\% of efficiency, this policy decreases the workload standard deviation by 78.6\%. Compared to the baselines, 
the trained policies achieve both better picking times and fairer workload distributions. 
Overall, the front pushes the boundaries of the pure performance and fairness policies, indicating that better trade-offs are hard to achieve. 
The similar conclusions can be found for warehouse types $M$ and $L$.
These results can provide decision-makers with several potential policies based on their preferences.

\begin{table}[htb]
\setlength{\tabcolsep}{0.5pt}
\scriptsize
    \begin{subtable}[t]{0.33\textwidth}
    \resizebox{\textwidth}{!}{%
    \centering
    \begin{tabular}{lcc}
    \toprule
    Policy & Picking Time & Workload SD \\ \midrule
    S1 & $15555 \pm 125$ & $41 \pm 4$ \\
    S2 & $12431 \pm 86$ & $43 \pm 4$ \\
    S3 & $9164 \pm 60$ & $66 \pm4$ \\
    S4 & $9188\pm 55$ & $114\pm 8$ \\
    S5 & $9074\pm 60$ & $118\pm 7$ \\
    S6 & $9149\pm 68$ & $167\pm 9$ \\
    Efficiency & $8586 \pm 62$ & $308 \pm 17$ \\
    Fairness & $19962 \pm 86$ & $61\pm9$ \\
    Greedy & $10619 \pm 59$ & $278 \pm 15$ \\
    VI Benchmark & $10087 \pm 58$ &  $442 \pm 23$\\ \bottomrule
    \end{tabular}}
    \caption{Warehouse type $S$.}
    \label{tab: MORL warehouse S}
    \end{subtable}\hfill
    \begin{subtable}[t]{0.33\textwidth}
    \resizebox{\textwidth}{!}{%
    \centering
    \begin{tabular}{lcc}
    \toprule
    Policy & Picking Time & Workload SD \\ \midrule
    M1 & $ 22180\pm65 $ & $ 86\pm 10 $ \\
    M2 & $ 18695\pm174 $ & $ 100\pm10 $ \\
    M3 & $ 14854\pm 74$ & $ 103\pm6$ \\
    M4 & $14897\pm 153 $ & $140\pm 9$ \\
    M5 & $9809\pm 169$ & $154\pm 8$ \\
    M6 & $9323\pm 136$ & $223\pm 11$ \\
    M7 & $8919\pm 51$ & $266\pm 19$ \\
    M8 & $8733\pm 52$ & $460\pm 32$ \\
    Efficiency & $8425\pm 46$  & $302\pm 13$ \\
    Fairness & $21793\pm 73$ & $73\pm4$ \\
    Greedy & $11023\pm 58 $ & $ 288\pm 9$ \\
    VI Benchmark & $10669 \pm 41$ &  $ 548\pm 17$\\ \bottomrule
    \end{tabular}}
    \caption{Warehouse type $M$.}
    \label{tab: MORL warehouse M}
    \end{subtable}\hfill
    \begin{subtable}[t]{0.33\textwidth}
    \resizebox{\textwidth}{!}{%
    \centering
    \begin{tabular}{lcc}
    \toprule
    Policy & Picking Time & Workload SD \\ \midrule
    L1 & $ 25562\pm 92$ & $ 70\pm 7$ \\
    L2 & $ 15474\pm 62 $ & $ 65\pm 3$ \\
    L3 & $8463\pm 32 $ & $ 72\pm 5$ \\
    L4 & $8296\pm 78$ & $76\pm 4$ \\
    L5 & $8116\pm 62$ & $139\pm 6$ \\
    L6 & $7400\pm 220$ & $226\pm 9$ \\
    Efficiency & $6540\pm37$  & $228\pm7$ \\
    Fairness & $21525 \pm 73$ & $51\pm3$ \\
    Greedy & $9823\pm 33$ & $ 253\pm 7$ \\
    VI Benchmark & $ 9569\pm61 $ &  $472 \pm 14$\\ \bottomrule
    \end{tabular}}
    \caption{Warehouse type $L$.}
    \label{tab: MORL warehouse L}
    \end{subtable}
    \caption[Performance of the non-dominated set of multi-objective policies learned on different warehouse types.]{Performance of the non-dominated set of policies learned on different warehouse types. The picking time is the average number of seconds to complete an episode, and the workload SD is the average standard deviation of the picker workloads in kilograms over 100 evaluation episodes. The $\pm$ indicates the 95\%-confidence interval.}
\label{tab: MORL performance evaluation}
\end{table}

\subsubsection{Policy Transferability to Various Picker/AMR Ratios}
To test how the learned policies perform in different resource situations, 
we use the policies trained in the performance evaluation experiment and evaluate each of these policies on 100 evaluation episodes for different picker/AMR ratios than they are trained on.
We test warehouse sizes with different picker/AMR ratios for warehouse types $S$ and $L$.
\begin{figure}[t]
\begin{subfigure}{0.315\textwidth}
   \centering
   \includegraphics[width=\linewidth]{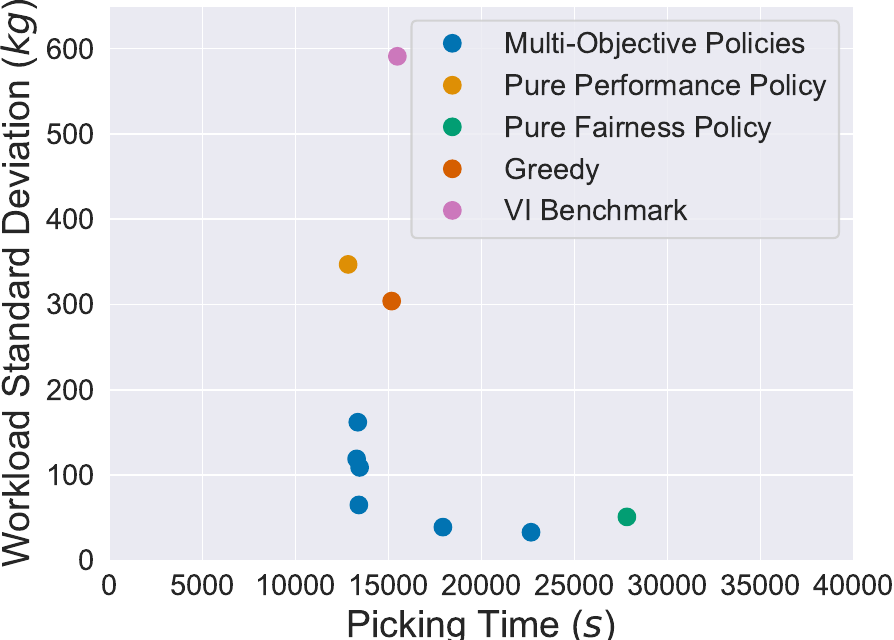}
   \caption{7 pickers and 15 \acrshortpl{amr}.}
   \label{fig: picker-amr-trans multi-obj S 7 15}
\end{subfigure}\hfill
\begin{subfigure}{0.3\textwidth}
   \centering
   \includegraphics[width=\linewidth]{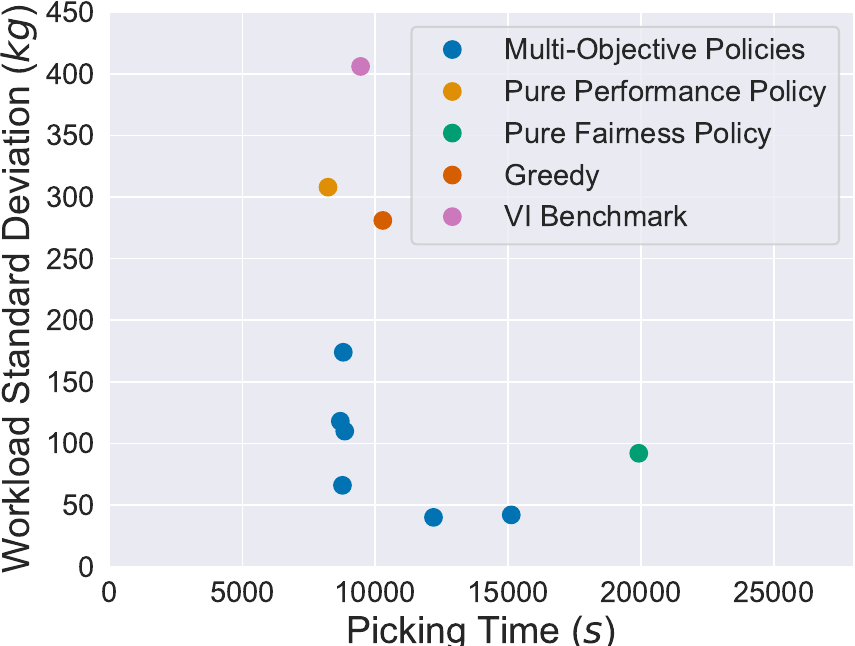}
   \caption{10 pickers and 30 \acrshortpl{amr}.}
   \label{fig: picker-amr-trans multi-obj S 10 30}
\end{subfigure}\hfill
\begin{subfigure}{0.315\textwidth}
   \centering
   \includegraphics[width=\linewidth]{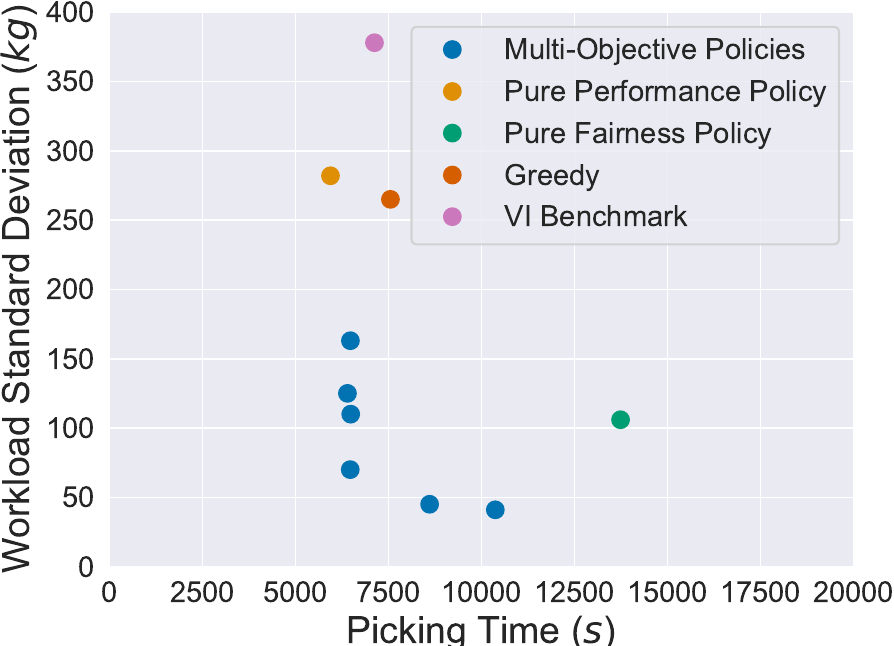}
   \caption{15 pickers and 35 \acrshortpl{amr}.}
   \label{fig: picker-amr-trans multi-obj S 15 35}
\end{subfigure}
\caption{Performance of the policies learned on warehouse type S when evaluated on different picker/AMR numbers.}
\label{fig: picker-amr-trans S}
\end{figure}

\begin{figure}[t]
\begin{subfigure}{0.315\textwidth}
    \centering
\includegraphics[width=\linewidth]{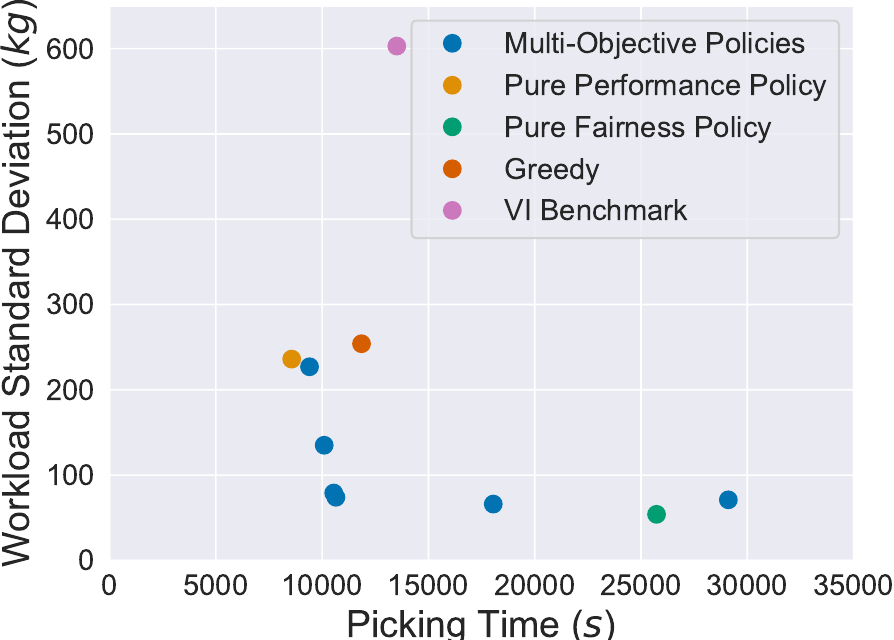}
    \caption{25 pickers and 60 \acrshortpl{amr}.}
    \label{fig: picker-amr-trans multi-obj L 25 60}
\end{subfigure}\hfill
\begin{subfigure}{0.3\textwidth}
    \centering
\includegraphics[width=\linewidth]{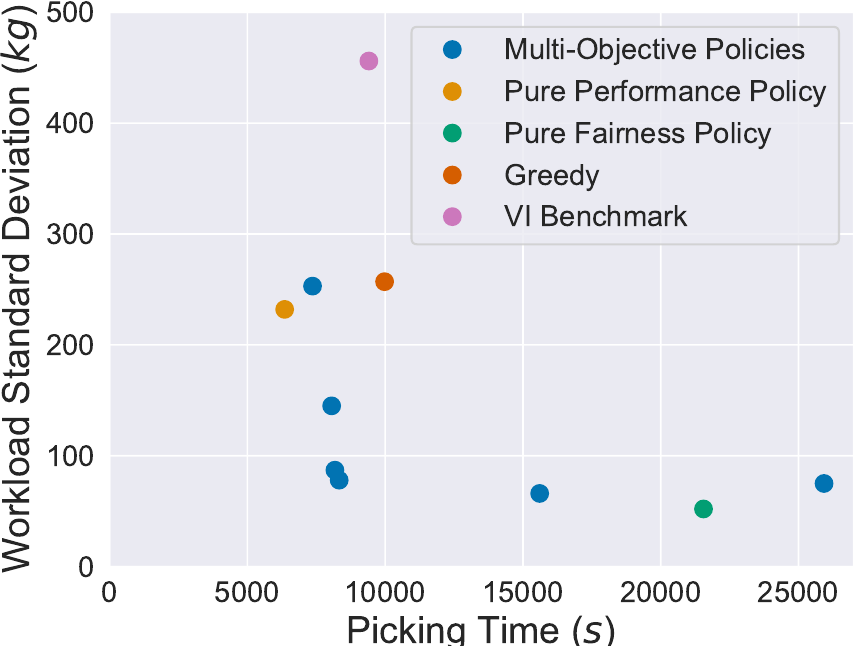}
    \caption{30 pickers and 100 \acrshortpl{amr}.}
    \label{fig: picker-amr-trans multi-obj L 30 100}
\end{subfigure}\hfill
\begin{subfigure}{0.3\textwidth}
    \centering
\includegraphics[width=\linewidth]{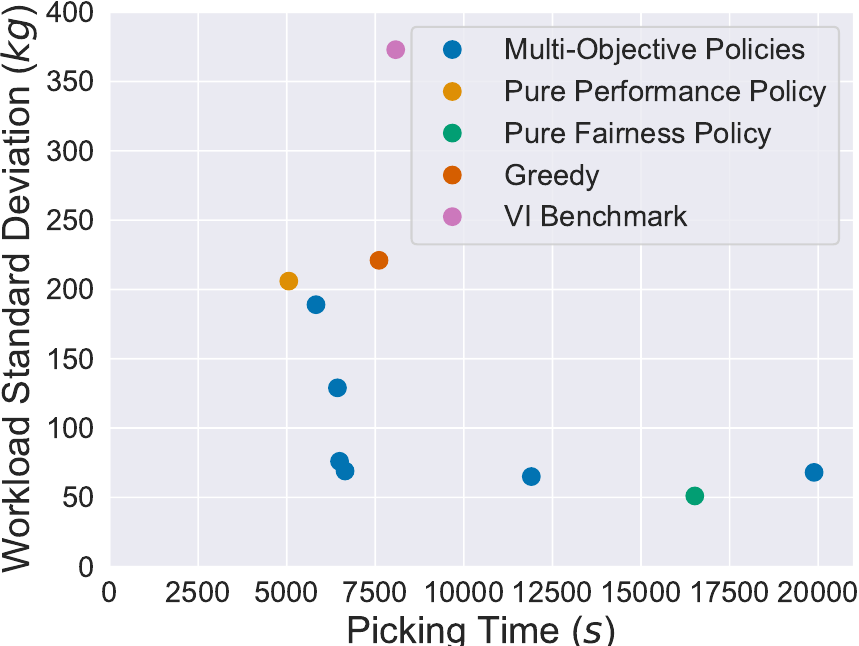}
    \caption{40 pickers and 110 \acrshortpl{amr}.}
    \label{fig: picker-amr-trans multi-obj L 40 110}
\end{subfigure}
\caption{Performance of the policies learned on warehouse type $L$ when evaluated on different picker/AMR numbers.}
\label{fig: picker-amr-trans L}
\end{figure}
Figures \ref{fig: picker-amr-trans S} and \ref{fig: picker-amr-trans L} show the multi-objective policies perform well in the different settings, as the policy front reaches similar levels compared to the pure efficiency and fairness policies.
The relative comparison between the policies looks like the front on the fixed evaluation warehouse. In this case, 
the fairness levels stay consistent for the different picker/\gls{amr} combinations. 
The pure fairness policy also maintains its fairness level with larger numbers of entities. In accordance with the previous results, for each combination of pickers and \glspl{amr}, several policies achieve better efficiency and fairness than the VI benchmark and greedy baseline.
For example, 
policy L4 achieves pick times  
and workload SD of 10545 and 79,
8177 and 87,
and 6491 and 76, respectively. These results are 22.0\%
13.1\%,
and 19.6\% better in terms of picking time and 86.9\%
80.9\%,
and 79.6\% better in terms of workload distribution than the VI benchmark. 

\subsubsection{Policy Transferability to Various Warehouse Sizes}
To show how the trained policies on fixed warehouse sizes perform on different sizes, 
we test the policies for types $S$, $M$, and $L$ on different warehouse sizes. 
We report evaluations on sizes 
$M$, $L$, and $XL$.
\begin{figure}
\centering
    \begin{subfigure}{0.3\textwidth}
        \centering
        \includegraphics[width=\linewidth]{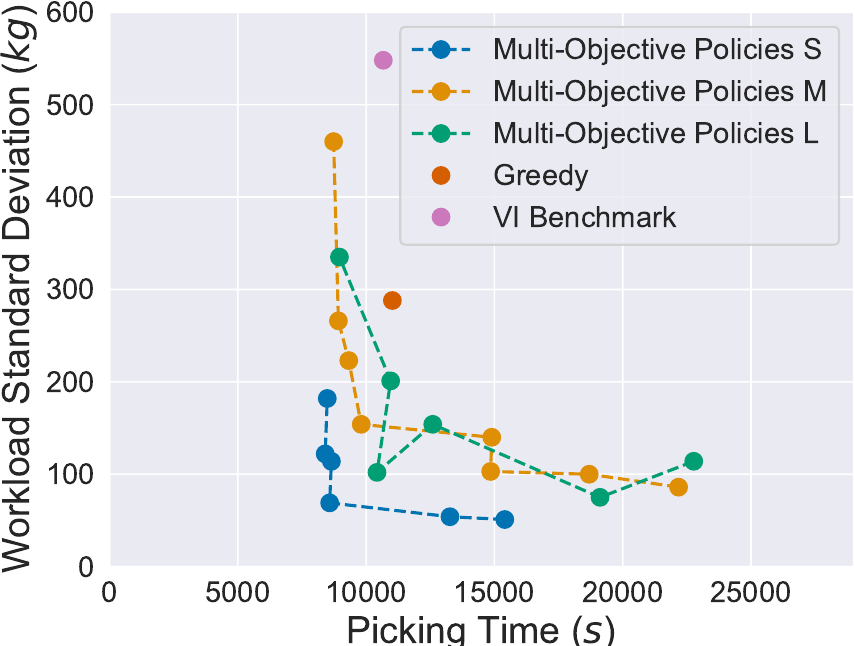}
        \caption{Evaluation on size $M$.}
        \label{fig: morl scalability eval M}
    \end{subfigure}\hfill
    \begin{subfigure}{0.3\textwidth}
        \centering
        \includegraphics[width=\linewidth]{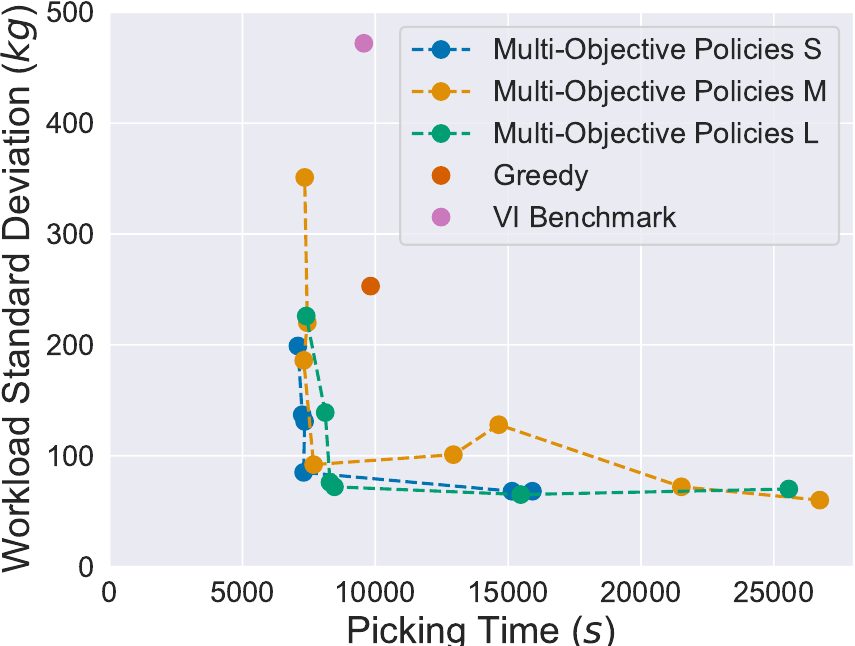}
        \caption{Evaluation on size $L$.}
        \label{fig: morl scalability eval L}
    \end{subfigure} \hfill
    \begin{subfigure}{0.3\textwidth}
        \centering
    \includegraphics[width=\linewidth]{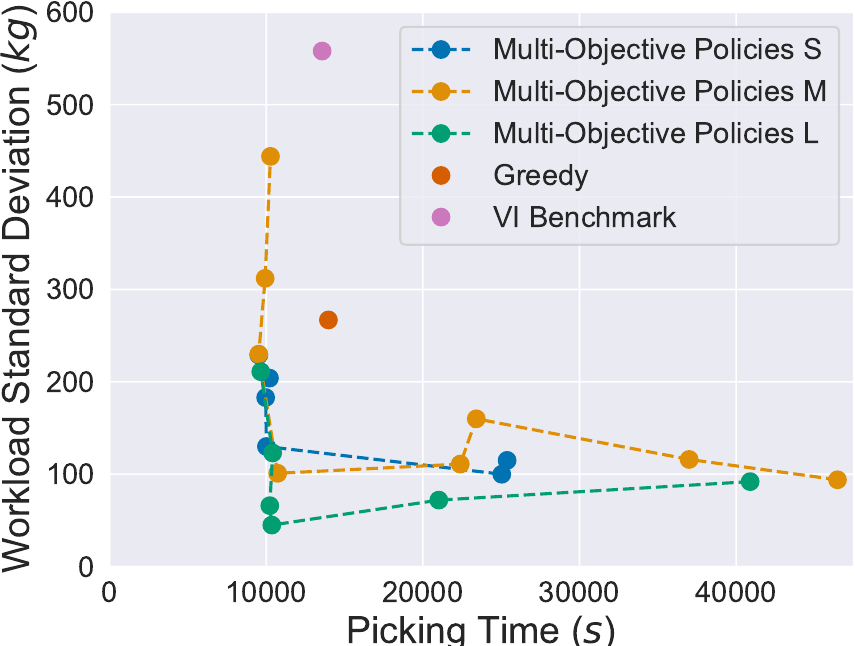}
        \caption{Evaluation size $XL$.}
        \label{fig: morl scalability eval XL}
    \end{subfigure}
    \caption{Performance of policies trained on different warehouse sizes when evaluated on varying warehouse sizes.}
    \label{fig: morl scalability eval}
\end{figure} 
Figure \ref{fig: morl scalability eval M} shows that for warehouse $M$, the type $S$ policies transfer remarkably well.
We find that all type $M$ policies are dominated by the type $S$ policies while evaluating for type $M$. Using the type $S$ policies, better combinations of fairness and efficiency are achieved than using the type $M$ policies. For example, policy S3 achieves an average completion time of 8578 seconds and workload standard deviation
of $69$ kg,
19.6\% and 87.4\% better than the VI benchmark.
The type $L$ policies transfer reasonably to warehouse size $M$. Figure \ref{fig: morl scalability eval M} shows that the fronts pass through each other, with more type $M$ policies having low picking times. All three fronts have several policies improving upon the baselines for both efficiency and workload fairness.
For type $L$ warehouses (Figure \ref{fig: morl scalability eval L}), the policy sets trained on the three different warehouse types form similar result fronts, showing one objective can be improved a lot without sacrificing much on the other. 
All sets contained policies that outperformed the benchmarks. 
\par
The evaluation on the $XL$ warehouses shows a slightly different pattern (Figure \ref{fig: morl scalability eval XL}). Here, the policy sets trained on the three different warehouse types formed similar result fronts, indicating good transferability. What stands out is that the policies focussing more on fairness deteriorate in terms of fairness compared to the more efficient policies, especially for the policy set $L$. In contrast, the fairness scores are similar or slightly better for the smaller sizes. Thus, policies with a significant focus on fairness may scale less well to larger warehouses in some cases. However, in practice, these policies will not often be selected as they achieved just a marginal fairness improvement while having much worse performance. On the other hand, the policies with better efficiency scale relatively well to the largest warehouse sizes, with policy set $L$ achieving the best trade-offs. 
For example, one policy scores an average picking time of 10357 with a workload standard deviation of 45 kg, constituting improvements of 23.6\% and 91.9\% over the VI benchmark scores of 13570 seconds and 558 kg, respectively. 
\par
These results show the practicality of our approach. The policies trained on specific warehouse sizes and picker/AMR ratios can be used directly for other situations, especially, larger and busier warehouses. The numbers corresponding to all figures are presented in the supplementary material. There, we also outline the transferability of the single-objective policies, showing similar results.

\subsubsection{Ablation Study}
To show the effectiveness of our proposed architecture, we evaluate the performance of the aisle-embedding (AISLE-EMB) architecture, compared to invariant feed-forward (INV-FF), \gls{gin}, and \gls{gcn} networks on single-objective efficiency performance.
\par
Table \ref{tab: architecture comparison} demonstrates that our network performs best on all warehouse sizes. Oppositely, the \gls{gin} and \gls{gcn} structures both perform poorly compared to the aisle-embedding and invariant feed-forward networks. Especially for the two larger warehouses, the difference is clear. Thus, message passing networks cannot sufficiently extract useful regional information. Instead, the extra parameters introduce noise into the learning process, limiting their performance.
The difference with the invariant feed-forward network is smaller. Even though the aisle-embedding actor outperforms it on each warehouse type, the difference is within a few percent. This difference may be so slight because we use multiple node features that already describe regional information related to efficiency. Still, the aisle-embedding architecture increases performance for single-objective optimization.
\begin{table}[ht]
\centering
\small
\setlength{\tabcolsep}{2.5pt}
\begin{tabular}{lcccc}
\toprule
Warehouse & INV-FF & AISLE-EMB & GIN & GCN \\ \midrule
$S$ & $8689 \pm 58$ & $\mathbf{8586 \pm 62}$ & $8869 \pm 55$ & $11677 \pm 67$ \\
$M$ & $ 8628\pm 40$ & $\mathbf{8425\pm 46}$ & $14151 \pm 75$ & $13851 \pm 65$\\
$L$ & $6602 \pm 29$ & $\mathbf{6540 \pm 37}$ & $11723 \pm 76$ & $14419 \pm 88$ \\ \bottomrule
\end{tabular}
\caption{Average picking times in seconds over 100 evaluation episodes of policies with different architectures. 
The bold markings indicate the best performances.}
\label{tab: architecture comparison}
\end{table}
\par
We further compare our architecture on various weighted-sum objectives balancing efficiency and fairness to demonstrate the good performance of AEMO-Net compared to other architectures, for which we refer to the supplementary material. In fact, for these weighted-sum objectives, the advantage is larger. This is likely because spatial information related to fairness is less easily captured in the node features and thus there is more dependence on the network architecture.
\section{Conclusion}
We present \gls{drl}-Guided Picker Optimization, which is a multi-objective \gls{drl} approach to simultaneously optimize and balance efficiency and fairness in collaborative human-robot order picking. In contrast to most prior works focused solely on deterministic scenarios without regard for fairness, we frame this as a sequential decision making problem under uncertainty. 
Experiment results demonstrate that our approach can find non-dominated policy sets that outline good trade-offs between fairness and efficiency.
The proposed AEMO-Net architecture is shown to be effective in 
capturing regional information and information regarding two objectives.  
 Furthermore, the approach is practical, in the sense that the learned policies exhibit good transferability to varying operational conditions and warehouse sizes.
Given the compelling advantages of our approach for complex, real-world 
settings, our industrial partner is currently implementing our method. As future work, we will investigate how to further account for possible practical preferences and constraints to solve relevant matching problems.

\bibliography{references}

\appendix
\input{appendix}
\end{document}

%% file: appendix.tex
\section{Single-Objective Transferability}
\subsection{Policy Transferability to Various Picker/AMR Ratios}
Table \ref{tab: amr-picker transferability} shows the results of the transferability analysis of the single-objective \gls{drl} policies to the different picker and \glspl{amr} numbers. 
\begin{table}[ht!]
    \footnotesize
    \begin{subtable}[h]{\linewidth}
    \renewcommand{\arraystretch}{0.67}
    \centering
    \label{tab: amr-picker transferability warehouse S}
    \begin{tabular}{lcccrc}
    \toprule
     & \multicolumn{2}{c}{\acrshort{drl}} & \multicolumn{2}{c}{Greedy} & VI Benchmark \\ \cmidrule(l){2-6} 
    Pickers/\acrshortpl{amr} & Picking Time & $\%$ & Picking Time & $\%$ & Picking Time \\ \midrule
    7/15 & $ \mathbf{12825\pm83}$ & $\mathbf{17.1}$ & $15166 \pm 74$ & $2.0$ & $ 15472\pm 87$ \\
    10/20 & $\mathbf{9206\pm 51}$ & $\mathbf{19.4}$ & $11274\pm 69$ & $1.3$ & $11420\pm 56$ \\
    10/30 & $\mathbf{8221\pm 54}$ & $\mathbf{13.0}$ & $10283\pm 60$ & $-8.8$ & $9447\pm 52$ \\
    15/25 & $\mathbf{6737\pm 42}$ & $\mathbf{21.5}$ & $7994\pm 40$ & $6.9$ & $8583\pm36$ \\
    15/30 & $\mathbf{5930\pm 34}$ & $\mathbf{24.7}$ & $7804\pm55$ & $1.0$ & $7879\pm 46$ \\
    15/35 & $\mathbf{5938\pm 35 }$ & $\mathbf{16.6}$ & $7550\pm 44$ & $-6.0$ & $7121\pm 38$ \\ \bottomrule
    \end{tabular}
    \caption{Warehouse type $S$.} 
    \end{subtable}
    \begin{subtable}{\linewidth}
    \renewcommand{\arraystretch}{0.67}
    \centering
    \label{tab: amr-picker transferability warehouse M}
    \begin{tabular}{lcccrc}
    \toprule
     & \multicolumn{2}{c}{\acrshort{drl}} & \multicolumn{2}{c}{Greedy} & VI Benchmark \\ \cmidrule(l){2-6} 
    Pickers/\acrshortpl{amr} & Picking Time & $\%$ & Picking Time & $\%$ & Picking Time \\ \midrule
    15/35 & $\mathbf{11263 \pm 50}$ & $\mathbf{21.2}$ & $ 14240\pm 66$ & $0.4$ & $14297\pm 72$ \\
    20/40 & $\mathbf{9139 \pm 46}$ & $\mathbf{23.5}$ & $11331\pm 56$ & $5.2$ & $11952\pm 41$ \\
    20/60 & $\mathbf{7965 \pm 48}$ & $\mathbf{16.0}$ & $10569\pm 51$ & $-11.4$ & $9489\pm 53$ \\
    30/50 & $\mathbf{6795 \pm 34}$ & $\mathbf{26.2}$ & $8189\pm 46$ & $11.0$ & $9206\pm34$ \\
    30/60 & $\mathbf{6293 \pm 38}$ & $\mathbf{22.7}$ & $7789\pm31$ & $4.3$ & $8136\pm 50$ \\
    30/70 & $\mathbf{5944 \pm 37}$ & $\mathbf{20.6}$ & $7620\pm 29$ & $-1.7$ & $7490\pm 41$ \\ \bottomrule
    \end{tabular}
    \caption{Warehouse type $M$.}
    \end{subtable}
    \begin{subtable}[h]{\linewidth}
    \renewcommand{\arraystretch}{0.67}
    \centering
    \label{tab: amr-picker transferability warehouse L}
    \begin{tabular}{lcccrc}
    \toprule
     & \multicolumn{2}{c}{\acrshort{drl}} & \multicolumn{2}{c}{Greedy} & VI Benchmark \\ \cmidrule(l){2-6} 
    Pickers/\acrshortpl{amr} & Picking Time & $\%$ & Picking Time & $\%$ & Picking Time \\ \midrule
    25/60 & $ \mathbf{8566\pm 34}$ & $\mathbf{36.6}$ & $ 11852\pm 31 $ & $12.3$ & $13512\pm 65$ \\
    30/70 & $\mathbf{7209\pm 26}$ & $\mathbf{37.7}$ & $10120\pm 35 $ & $12.5$ & $11563\pm 58$ \\
    30/100 & $\mathbf{6354\pm 65}$ & $\mathbf{32.5}$ & $9980\pm 44$ & $-6.0$ & $9410\pm 62$ \\
    40/90 & $\mathbf{5659\pm 29}$ & $\mathbf{36.1}$ & $7962\pm 29$ & $10.1$ & $8859\pm68$ \\
    40/100 & $\mathbf{5279\pm 50}$ & $\mathbf{34.6}$ & $8141\pm44$ & $3.4$ & $8424\pm 52$ \\
    40/110 & $\mathbf{5059\pm 18}$ & $\mathbf{37.3}$ & $7605\pm 27$ & $5.8$ & $8076\pm 45$ \\ \bottomrule
    \end{tabular}
    \caption{Warehouse type $L$.} 
    \end{subtable}
    \begin{subtable}[h]{\linewidth}
    \renewcommand{\arraystretch}{0.67}
    \centering
    \label{tab: amr-picker transferability warehouse XL}
    \begin{tabular}{lcccrc}
    \toprule
     & \multicolumn{2}{c}{\acrshort{drl}} & \multicolumn{2}{c}{Greedy} & VI Benchmark \\ \cmidrule(l){2-6} 
    Pickers/\acrshortpl{amr} & Picking Time & $\%$ & Picking Time & $\%$ & Picking Time \\ \midrule
    50/120 & $\mathbf{12028\pm 23}$ & $\mathbf{40.2}$ & $16816 \pm 32 $ & $16.5$ & $20142\pm 112$ \\
    60/140 & $\mathbf{10150\pm 20}$ & $\mathbf{40.7}$ & $14312\pm  27$ & $16.4 $ & $17118\pm 101$ \\
    60/200 & $\mathbf{9009\pm 44}$ & $\mathbf{35.6}$ & $14293\pm 88$ & $-2.2 $ & $13979\pm 87$ \\
    80/180 & $\mathbf{8106\pm 77}$ & $\mathbf{38.9}$ & $11343\pm 30$ & $14.6 $ & $13275\pm83$ \\
    80/200 & $\mathbf{8011\pm 59}$ & $\mathbf{36.8 }$ & $11765\pm 52$ & $6.4 $ & $12571\pm 91$ \\
    80/220 & $\mathbf{6947\pm 19}$ & $\mathbf{41.5}$ & $10799 \pm 40$ & $ 9.1$ & $11877\pm 84$ \\ \bottomrule
    \end{tabular}
    \caption{Warehouse type $XL$.} 
    \end{subtable}
    \caption{Performance of \acrshort{drl} policies
    given varying picker/\gls{amr} combinations. The values indicate the picking time in seconds. The $\pm$ indicates the 95\%-confidence interval. The $\%$ indicates the percentage improvement over the VI Benchmark, with a positive percentage indicating an improvement and, thus, lower times. The bold markings indicate the best performance values per warehouse setting.}
    \label{tab: amr-picker transferability}
\end{table}

\par
The \gls{drl} approach outperforms both greedy and the VI Benchmark for each combination of pickers and \glspl{amr} in each warehouse size. The performance improvement over the VI Benchmark is the largest for the larger warehouses. 
Remarkably, the relative improvement of the picking times is better for most picker/\gls{amr} ratios than the improvements for the trained warehouse instances. On warehouse types $S$ and $M$, the advantage is only smaller for the ratios 10/30 and 20/60, respectively. These are ratios with a relatively low number of pickers and, in comparison, many \glspl{amr}. For all other combinations, the percentage improvement over the VI Benchmark is roughly equal or better. This shows that, whereas the VI Benchmark efficiency deteriorates when the crowdedness levels in the warehouse become either small or larger, the \gls{drl} policy continues to achieve good results. Thus, the \gls{drl} policy can adapt to extremer warehouse occupation levels more efficiently.
\par
In several cases, the greedy baseline performs slightly better than the VI Benchmark, with the best of the two alternating for different settings. Especially for the larger warehouse size with extremer picker/\gls{amr} numbers, the greedy policy seems more suitable. However, the greedy baseline, like the VI benchmark, does not get close to the \gls{drl} performance for any problem instance.

\subsection{Policy Transferability to Various Warehouse Sizes}
Table \ref{tab: warehouse size transferability} shows the results of the transferability analysis of the \gls{drl} policies to the different warehouse types. The results reveal that the policies adapt well to different warehouse sizes. We see that the policy trained on warehouse type $S$ achieves an average total pick time of 6877 seconds on type $L$ compared to the 6540 seconds reached by the policy trained on warehouse $L$. Thus, while being developed for a warehouse with over 6 times fewer pick locations and roughly 3 times as little pickers and \glspl{amr}, it only performs about 5\% worse. Similarly, the policies also scale down well to smaller warehouses. The policy of warehouse type $L$ achieves an average completion time of 8875 seconds compared to the 8586 seconds of policy $S$. This is a performance difference of just over 3\%. Remarkably, policy $L$ (8567 seconds) outperforms policy $XL$ (9010) on all instance sizes. Policy $L$ achieves an improvement of 36.9\% over the VI benchmark, compared to the 33.6\% improvement of policy $XL$. This indicates that training for increasingly larger warehouse sizes is not necessary to get good performance on those warehouse sizes. In larger warehouse sizes, the action space is bigger, and therefore, learning can be slower and harder to fine-tune to get the last percentage improvements. Learning for many more iterations might eventually bring better results, but this is not guaranteed and the learning is substantially slower, as we already train the $XL$ policy for twice as many steps as those for types $M$ and $L$. The $XL$ policy does transfer well to other warehouse sizes though, which again indicates the good transferability of policies. In addition to the comparative performances between each other, all \gls{drl} policies maintain a clear advantage over the greedy and VI benchmark results.
\par
Thus, overall, the policies adapt well to different warehouse sizes. This enables easier deployment of policies to varying warehouses. Also, when a warehouse layout is changed, the policies can maintain good performance without needing to retrain and redeploy new policies. In addition, it is advantageous for the training process itself since one can train and evaluate different settings quicker on smaller warehouse instances and then scale the learned policies to larger warehouses.

\begin{table}[t]
\renewcommand{\arraystretch}{0.67}
\centering
\resizebox{\textwidth}{!}{%
\begin{tabular}{lcccccc}
\toprule
Warehouse & Policy $S$ & Policy $M$ & Policy $L$ & Policy $XL$ &Greedy & VI Benchmark \\ \midrule
$S$ & $\mathbf{8586\pm 62}$ & $9190\pm 53$ & $8875\pm58$ & $8986 \pm 51 $ &$10619\pm59$ & $10087\pm58$\\
$M$ & $\mathbf{7931\pm 42}$ & $8425\pm46$ & $8064\pm41$ & $8220 \pm 37$&$11023\pm58$ & $10669\pm41$ \\
$L$ & $6877\pm31$ & $7190\pm42$  & $\mathbf{6540\pm37}$ & $ 6877 \pm 23$ &$9823\pm33$& $9569\pm61$ \\
$XL$ & $9478\pm 20$ & $11275\pm33$ & $\mathbf{8567\pm24}$ & $9010 \pm 21$&$13972\pm44$ &$13570\pm72$  \\ \bottomrule
\end{tabular}
}
\caption{Performance of \acrshort{drl} policies when evaluated on a variety of warehouse sizes. The values indicate the picking time in seconds. The $\pm$ indicates the 95\%-confidence interval. Policy $X$ indicates the \gls{drl} policy trained on warehouse type $X$. The bold markings indicate the best performance values per warehouse size.} \label{tab: warehouse size transferability}
\end{table}

\section{Multi-Objectives Experiments: Tables}
\subsection{Policy Transferability to Various Picker/AMR Ratios}
Table \ref{tab: pick-amr-trans Multi-objective} shows the detailed numerical results belonging to the multi-objective transferability experiments of different picker/\gls{amr} ratios.

\begin{table}[h!]
    
    \begin{subtable}{\linewidth}
    \renewcommand{\arraystretch}{0.67}
    \centering
    \setlength{\belowcaptionskip}{6pt}
    \begin{tabular}{lcccccc}
    \toprule
     & \multicolumn{2}{c}{7 Pickers/15 \acrshortpl{amr}} & \multicolumn{2}{c}{10 Pickers/30 \acrshortpl{amr}} & \multicolumn{2}{c}{15 Pickers/35 \acrshortpl{amr}} \\ \cmidrule(l){2-7} 
    Policy & PT & WF & PT & WF & PT & WF \\ \midrule
    S1 & $22659\pm 213$ & $33\pm 4$ & $15117\pm 133$ & $42\pm4$ & $10369\pm96$ & $41\pm3$ \\
    S2 & $17921\pm134$ & $39\pm4$ & $ 12191\pm 85 $ & $40\pm3$ & $8603\pm56 $ & $45\pm3$ \\
    S3 & $13402\pm87$ & $65\pm4$ & $8765\pm64$ & $66\pm5$ & $6469\pm 52$ & $70\pm4$ \\
    S4 & $13443\pm87$ & $109\pm8$ & $8850\pm66$ & $110\pm7$ & $6482\pm48$ & $110\pm6$ \\
    S5 & $13281\pm107$ & $119\pm10$ & $8684\pm70$ & $118\pm8$ & $6394\pm65$ & $125\pm6$ \\
    S6 & $13345\pm113$ & $162\pm12$ & $8795\pm95$ & $174\pm13$ & $6474\pm72$ & $163\pm11$ \\
    Pure Performance & $12825\pm 83$ & $347 \pm 23$ & $8221\pm 54$ & $308 \pm 20$ & $5938\pm 35$ & $282 \pm 15$\\
    Pure Fairness & $27812 \pm 115$ & $51 \pm 5$ & $19916 \pm 104$ & $92 \pm 12$ & $13736 \pm 73$ & $106 \pm 12$ \\
    Greedy & $15166 \pm 74$ & $304 \pm 21$& $10283\pm 60$ & $281 \pm 15$ & $7550\pm 44$ & $265 \pm 11 $\\
    VI Benchmark & $ 15472\pm 87$ & $591 \pm 40$ & $9447\pm 52$ & $406 \pm 22$ & $7121\pm 38$ & $378 \pm 15$\\
    \bottomrule
    \end{tabular}
    \caption{Warehouse type $S$.}
    \label{tab: pick-amr-trans Multi-obj warehouse S}
    \end{subtable}

    \begin{subtable}{\linewidth}
    \renewcommand{\arraystretch}{0.67}
    \centering
    \begin{tabular}{lcccccc}
    \toprule
     & \multicolumn{2}{c}{25 Pickers/60 \acrshortpl{amr}} & \multicolumn{2}{c}{30 Pickers/100 \acrshortpl{amr}} & \multicolumn{2}{c}{40 Pickers/110 \acrshortpl{amr}} \\ \cmidrule(l){2-7} 
    Policy & PT & WF & PT & WF & PT & WF \\ \midrule
    L1 & $29109\pm 82$ & $71\pm 6$ & $25928\pm 93$ & $75\pm7$ & $19885\pm75$ & $68\pm4$ \\
    L2 & $18050\pm 62$ & $66\pm3$ & $15608\pm 53$ & $66\pm3$ & $11904\pm 48$ & $65\pm2$ \\
    L3 & $10647\pm78$ & $74\pm6$ & $8332\pm64$ & $78\pm6$ & $6647\pm 63$ & $69\pm5$ \\
    L4 & $10545\pm81$ & $79\pm4$ & $8177\pm76$ & $87\pm5$ & $6491\pm68$ & $76\pm5$ \\
    L5 & $10095\pm 91$ & $135\pm6$ & $8059\pm73$ & $145\pm6$ & $6432\pm62$ & $129\pm5$ \\
    L6 & $9407\pm 42$ & $227\pm10$ & $7365\pm61$ & $253\pm11$ & $5826\pm56$ & $189\pm7$\\
    Pure Performance &$ 8566\pm 34$ & $236 \pm 7$& $6354\pm 65$& $232 \pm 7$& $5059\pm 18$& $206 \pm 5$\\
    Pure Fairness & $25731 \pm 71$ & $54 \pm 5$ & $21554 \pm 72$ & $52 \pm 3$ & $16518 \pm 64$ & $51 \pm 3$\\
    Greedy & $11852\pm 31$ & $254 \pm 8$& $9980\pm 44$& $257 \pm 7$ & $7605\pm 27$ & $221 \pm 6$\\
    VI Benchmark & $13512\pm 65$ & $603 \pm 15$& $9410\pm 62$ & $456 \pm 13$ & $8076\pm 45$& $373 \pm 9$ \\
    \bottomrule
    \end{tabular}
    \caption{Warehouse type $L$.}
    \label{tab: pick-amr-trans Multi-obj warehouse L}
    \end{subtable}
    \caption[Performance of multi-objective DRL policies trained on warehouse type $S$ and $L$, given varying combinations of the number of pickers and AMRs within their respective warehouse sizes.]{Performance of multi-objective \acrshort{drl} policies trained on warehouse type $S$ and $L$, given varying combinations of the number of pickers and \acrshortpl{amr} within their respective warehouse sizes. PT is the picking time in seconds and WF is the standard deviation of the workloads in kilograms. The $\pm$ indicates the 95\%-confidence interval.}
    \label{tab: pick-amr-trans Multi-objective}
\end{table}

\subsection{Policy Transferability to Various Warehouse Sizes}
Table \ref{tab: transferability size multi-objective} shows the detailed numerical results belonging to the multi-objective transferability experiments of different warehouse sizes.
\begin{table}[p]
    \begin{subtable}{\linewidth}
    \renewcommand{\arraystretch}{0.67}
    \centering
    \begin{tabular}{lcccccc}
    \toprule
     & \multicolumn{2}{c}{$S$} & \multicolumn{2}{c}{$M$} & \multicolumn{2}{c}{$L$} \\ \cmidrule(l){2-7} 
    Policy nr. & PT & WF & PT & WF & PT & WF \\ \midrule
    1 & $15555 \pm 125$ & $41 \pm 4$ & $20876 \pm 129$ & $95 \pm 19$ & $21182 \pm 107$ & $96 \pm 14$ \\
    2 & $12431 \pm 86$ & $43 \pm 4$ & $18938 \pm 146$ & $98 \pm 10$ & $19888 \pm 95$ & $82 \pm 86$ \\
    3 & $9164 \pm 60$ & $66 \pm4$ & $14666 \pm 96$ & $74 \pm 5$ & $13516 \pm 140$ & $112 \pm 12$ \\
    4 & $9188\pm 55$ & $114\pm 8$ & $14879 \pm 134$ & $106 \pm 11$ & $12163 \pm 144$ & $81 \pm 89$ \\
    5 & $9074\pm 60$ & $118\pm 7$& $11193 \pm 147$ & $155 \pm 11$ & $11621 \pm 147$ & $200 \pm 15$  \\
    6 & $9149\pm 68$ & $167\pm 9$ & $10302 \pm 168$ & $226 \pm 15$ & $10303 \pm 113$ & $355 \pm 28$ \\
    7 & - & - & $9577 \pm 53$ & $206 \pm 18$ & - & - \\
    8 & - & - & $9464 \pm 57$ & $441 \pm 51$ & - & - \\ \bottomrule
    \end{tabular}
    \caption{Evaluation results on warehouse type $S$.}
    \label{tab: transferability size multi-objective type S}
    \end{subtable}

    \begin{subtable}{\linewidth}
    \renewcommand{\arraystretch}{0.67}
    \centering
    \begin{tabular}{lcccccc}
    \toprule
     & \multicolumn{2}{c}{$S$} & \multicolumn{2}{c}{$M$} & \multicolumn{2}{c}{$L$} \\ \cmidrule(l){2-7} 
    Policy nr. & PT & WF & PT & WF & PT & WF \\ \midrule
    1 & $15404 \pm 100$ & $ 51\pm 4$ & $ 22180\pm65 $ & $ 86\pm 10 $ & $22770 \pm 102$ & $114 \pm 10$ \\
    2 & $13267 \pm 63$ & $ 54\pm 5$ & $ 18695\pm174 $ & $ 100\pm10 $ & $ 19117\pm 77$ & $75 \pm 3$ \\
    3 & $ 8578\pm 69$ & $ 69\pm 4$ & $ 14854\pm 74$ & $ 103\pm6$ & $ 12596\pm 177$ & $154 \pm 9$ \\
    4 & $8646\pm 49$ & $114\pm 5$ & $14897\pm 153 $ & $140\pm 9$ & $10424 \pm 96$ & $ 102\pm 9$ \\
    5 & $8405\pm 50$ & $122\pm 6$& $9809\pm 169$ & $154\pm 8$ & $10956 \pm 185$ & $201 \pm 10$  \\
    6 & $8485\pm 63$ & $182\pm 9$ & $9323\pm 136$ & $223\pm 11$ & $ 8960\pm 71$ & $ 335\pm 22$ \\
    7 & - & - & $8919\pm 51$ & $266\pm 19$ & - & - \\
    8 & - & - & $8733\pm 52$ & $460\pm 32$ & - & - \\ \bottomrule
    \end{tabular}
    \caption{Evaluation results on warehouse type $M$.}
    \label{tab: transferability size multi-objective type M}
    \end{subtable}

    \begin{subtable}{\linewidth}
    \renewcommand{\arraystretch}{0.67}
    \centering
    \begin{tabular}{lcccccc}
    \toprule
     & \multicolumn{2}{c}{$S$} & \multicolumn{2}{c}{$M$} & \multicolumn{2}{c}{$L$} \\ \cmidrule(l){2-7} 
    Policy nr. & PT & WF & PT & WF & PT & WF \\ \midrule
    1 & $ 15913\pm 90$ & $ 68\pm 7$ & $ 26728 \pm 159$ & $ 60\pm 5 $ & $ 25562\pm 92$ & $ 70\pm 7$ \\
    2 & $15146 \pm 67$ & $ 68\pm 7$ & $ 21520\pm 97$ & $ 72\pm 6$ & $ 15474\pm 62 $ & $ 65\pm 3$ \\
    3 & $7302 \pm 62$ & $85 \pm 6$ & $ 14645 \pm 70$ & $ 128\pm 7$ & $8463\pm 32 $ & $ 72\pm 5$ \\
    4 & $7340\pm 53$ & $131\pm 6$ & $12939\pm 81 $ & $101\pm 5$ & $8296\pm 78$ & $76\pm 4$ \\
    5 & $7249\pm 137$ & $137\pm 6$& $7678\pm 71$ & $92\pm 6$ & $8116\pm 62$ & $139\pm 6$  \\
    6 & $7087\pm 64$ & $199\pm 8$ & $7307\pm 80$ & $ 186 \pm 8$ & $7400\pm 220$ & $226\pm 9$ \\
    7 & - & - & $7442\pm 42$ & $220\pm 12$ & - & - \\
    8 & - & - & $7343\pm 43$ & $351\pm 18$ & - & - \\ \bottomrule
    \end{tabular}
    \caption{Evaluation results on warehouse type $L$.}
    \label{tab: transferability size multi-objective type L}
    \end{subtable}
\end{table}
\begin{table} \ContinuedFloat
    \begin{subtable}{\linewidth}
    \renewcommand{\arraystretch}{0.67}
    \centering
    \begin{tabular}{lcccccc}
    \toprule
     & \multicolumn{2}{c}{$S$} & \multicolumn{2}{c}{$M$} & \multicolumn{2}{c}{$L$} \\ \cmidrule(l){2-7} 
    Policy nr. & PT & WF & PT & WF & PT & WF \\ \midrule
    1 & $ 25380\pm 81$ & $115\pm 9$ & $ 46473 \pm337$ & $94 \pm 5 $ & $40897 \pm 125$ & $92 \pm 5$ \\
    2 & $ 25039\pm 72$ & $100\pm 10$ & $37004 \pm 231$ & $116\pm 5$ & $ 21019\pm 67 $ & $ 72\pm 2$ \\
    3 & $ 10013\pm 108$ & $130 \pm 6$ & $  23413\pm 65$ & $ 160\pm 5$ & $10357\pm 94 $ & $45 \pm 4$ \\
    4 & $9964\pm 87$ & $183\pm 8$ & $22389\pm 267$ & $111\pm 4$ & $10229\pm 112$ & $66 \pm 4$\\
    5 & $10208\pm 83$ & $204\pm 8$& $10730\pm 151$ & $101\pm 5$ & $10411\pm 91$ & $123 \pm 4$\\
    6 & $9528\pm 42$ & $229\pm 6$ & $ 9524 \pm 96$ & $230\pm 7$ & $9653\pm 30$ & $211 \pm 7$\\
    7 & - & - & $9932\pm 92$ & $312\pm 3$ & - & - \\
    8 & - & - & $10173\pm 113$ & $444\pm 17$ & - & - \\ \bottomrule
    \end{tabular}
    \caption{Evaluation results on warehouse type $XL$.}
    \label{tab: transferability size multi-objective type XL}
    \end{subtable}
    \caption[Performance of multi-objective DRL policies when evaluated on various warehouse sizes.]{Performance of multi-objective \acrshort{drl} policies when evaluated on various warehouse sizes. PT is the picking time in seconds and the workload fairness WF is the standard deviation of the workloads in kg. The $\pm$ indicates the 95\%-confidence intervals. $S$, $M$, and $L$ in the columns indicate the training warehouse types of the policies.}
\label{tab: transferability size multi-objective}
\end{table}

\section{Ablation Study}
\subsection{Additional Experiment}
To further evaluate the performance of our architecture, we test the performance of AEMO-Net compared to just an aisle-embedding (AISLE-EMB), an invariant feed-forward network with feature separation (INV-FF-SEP), and a regular invariant feed-forward network (INV-FF) for several weight vectors leading to different weighted-sum rewards.
\par
Table \ref{tab: weight-sum comparison} outlines these results. The first thing that stands out is the performance difference between the aisle-embedding architectures and the invariant feed-forward architectures. On 5 of the 6 settings, the aisle-embedding instances achieve better rewards than the invariant feed-forward policies by a clear margin. Thus, whereas with single-objective optimization the differences between aisle-embedding and invariant feed-forward actors are small, the differences are more prominent when both fairness and performance must be optimized. A possible explanation is that the node features can capture less regional information regarding fairness. Hence, the aisle-embedding architecture has more possibilities to aid in extracting relevant regional information from the graph.
\begin{table}[t]
   \centering
   \scriptsize
   \setlength{\tabcolsep}{1.5pt}
    \resizebox{\linewidth}{!}{
   \begin{tabular}{lcccccccccccccc}
   \toprule
    &  &  & \multicolumn{3}{c}{AISLE-EMB-SEP} & \multicolumn{3}{c}{AISLE-EMB} & \multicolumn{3}{c}{INV-FF-SEP} & \multicolumn{3}{c}{INV-FF} \\ \cmidrule(l){4-15}
   Type & $w_{perf}$ & $w_{fair}$ & Reward & PT & WF & Reward & PT & WF & Reward & PT & WF & Reward & PT & WF \\
   \midrule
   $S$ & 0.5 & 0.5 & $-264 \pm 7 $ & $17017 \pm 180 $ & $100 \pm 11$ & $\mathbf{-231 \pm 6}$ & $\mathbf{14693 \pm 209}$ & $\mathbf{91 \pm 11}$ & $-372 \pm 5$ & $24400 \pm 83$ & $129 \pm 9$ & $-543 \pm 12$ & $22410 \pm 112$ & $506 \pm 26$ \\
   $S$ & 0.9 & 0.1 & $\mathbf{-236 \pm 4}$ & $\mathbf{10210 \pm 182}$ & $\mathbf{72 \pm 10}$  & $-379 \pm 3$ & $16366 \pm 148$ & $110 \pm 9$ & $-288 \pm 2$ & $12055 \pm 74$ & $166 \pm 10$ & $-536 \pm 3$ & $21507 \pm 118$ & $502 \pm 25$ \\
   $S$ & 0.1 & 0.9 & $-149 \pm 11$ & $21265 \pm 113$ & $102 \pm 11$ & $\mathbf{-138 \pm 9}$ & $\mathbf{22106 \pm 102}$ & $\mathbf{89 \pm 9}$ & $-187 \pm 13$ & $21808 \pm 126$ & $142 \pm 14$ & $-179 \pm 11$ & $21937 \pm 88$ & $133 \pm 12$ \\
   $M$ & 0.5 & 0.5 & $-339 \pm 5$ & $20340 \pm 141$ & $166 \pm 10$ & $-384 \pm 5$ & $23555 \pm 100$ & $175 \pm 10$ & $\mathbf{-255 \pm 7}$ & $\mathbf{17023 \pm 90}$ & $\mathbf{230 \pm 8}$ & $-381 \pm 6$ & $20989 \pm 110$ & $232 \pm 11$ \\
   $M$ & 0.9 & 0.1 & $\mathbf{-361 \pm 3}$ & $\mathbf{15305 \pm 139}$ & $\mathbf{163 \pm 8}$ & $-463 \pm 2$ & $20154 \pm 80$ & $100 \pm 6$ & $-463 \pm 3$ & $19324 \pm 100$ & $282 \pm 12$ & $-358 \pm3$ & $15102 \pm 74$  & $186 \pm 9$ \\
   $M$ & 0.1 & 0.9 & $-202 \pm 9$ & $25250 \pm 84$ & $151 \pm 9$ & $\mathbf{-197 \pm 7}$ & $\mathbf{23302 \pm 86}$ & $\mathbf{151 \pm 8}$ & $-253 \pm 7$ & $27338 \pm 95$ & $200 \pm 8$ & $-256 \pm 7$ & $16946 \pm 93$  & $232 \pm 6$  \\
    \bottomrule
   \end{tabular}%
    }
   \caption{Performance comparison of policies with different network architectures, trained using a weighted-sum reward between performance and fairness for various warehouse sizes and weight combinations for performance ($w_{perf}$) and fairness ($w_{fair}$). The table shows the obtained reward, the total picking time in seconds (PT), and the standard deviation of the picker workloads in kg (WF). $\pm$ indicates the 95\%-confidence interval. The bold markings indicate the policies with the best rewards per scenario.}
   \label{tab: weight-sum comparison}
\end{table}
\par
In addition, the results show that the aisle-embedding without feature separation reaches slightly better rewards than the actor with feature separation in three instances. However, the improvements are only marginal. Namely, for the weight vector $(0.1, 0.9)$, the final rewards of the two structures are very close and within each other's 95\%-confidence interval, indicating that we cannot conclude a statistically significant difference. Additionally, the reward difference is relatively small for weight vector $(0.5, 0.5)$ on warehouse $S$. Oppositely, in the cases in which the actor with feature separation performs better, the difference in rewards is much larger, being $-236$ versus $-379$ and $-361$ versus $-463$. Moreover, we observe that in these instances, with these weight vectors, the best overall policies for the warehouse types are found. Namely, both policies dominate all other policies for all weight vectors on both picking time and workload fairness. Thus, this weight vector region in which the aisle-embedding with feature separation outperforms the other architectures is also the region where the best policies are achievable. Hence, the aisle-embedding structure with feature separation is the best network architecture for the multi-objective learning task.
\par
What is also noteworthy in these results is that it is hard to judge which weight vectors lead to which trade-offs between efficiency and fairness. For example, we find that the policies for weights $w_{perf}=0.9$ and $w_{fair}=0.1$ score very well on both efficiency and fairness and even achieve better fairness than the policies with $w_{perf}=0.1$ and $w_{fair}=0.9$. In addition, the outcome of different weight settings varies between warehouse sizes. For example, for the aisle-embedding without feature separation, the best efficiency and fairness scores in warehouse type $S$ are achieved for weight vector $(0.5,0.5)$, whereas for type $M$ the policy for weight vector $(0.9, 0.1)$ outperforms the other policies. These findings highlight the value of using a multi-objective learning algorithm to find the weights that form a high-quality non-dominated set of policies. Otherwise, trying to hand-tune the weights for each problem instance would cost a vast amount of computational resources and effort to find clear trade-off fronts.

\subsection{Network Architectures}
\subsubsection{Invariant Feed-Forward Network}
For the invariant feed-forward actor network, we used two fully-connected layers with Leaky ReLU activation and 64 neurons, followed by a fully-connected layer with 16 neurons and Leaky ReLU and a last layer with one neuron that represents the node value, which is masked and passed through the softmax function with all nodes.
\par
In the critic network, we used three fully-connected layers with the Leaky ReLU activation function to create the node-embeddings. For the first two layers, we used 64 neurons, while the third layer had 16 neurons. Then, after aggregating the node-embeddings to form the graph-embedding, we used one fully-connected layer with one neuron to get the value estimate.

\subsubsection{Graph Networks}
For the GCN actor, we used four consecutive GCN layers with 64 output channels and Leaky ReLU activation function, followed by two fully-connected feed-forward layers of 64 and 16 neurons with Leaky ReLU, and a last fully-connected layer with one neuron. The GCN critic also had four consecutive GCN layers with 64 output channels and Leaky ReLU activation function, followed by two fully-connected feed-forward layers of 64 and 16 neurons with Leaky ReLU. These were followed by the summation aggregation and one final linear layer with one neuron to output the graph value.
\par
The GIN networks had the same structure as the GCN networks with GIN layers instead of GCN layers. For each GIN layer, we used a multilayer perceptron with two fully-connected layers of 64 neurons with Leaky ReLU activation.

\subsubsection{Other Networks} 
For the AISLE-EMB, we used the same structure used for the single-objective pure efficiency and pure fairness actors. For the INV-FF-SEP, we used an invariant feed-forward structure to create the efficiency and workload fairness embeddings. This invariant feed-forward structure consisted of two fully-connected layers with 64 neurons and a Leaky ReLU activation function followed by one fully-connected layer with 16 neurons. 

\section{Schematic Overviews Simulation Model}
\subsection{Picker Process}
\begin{figure}[th]
    \centering
    \includegraphics[width=0.5\linewidth]{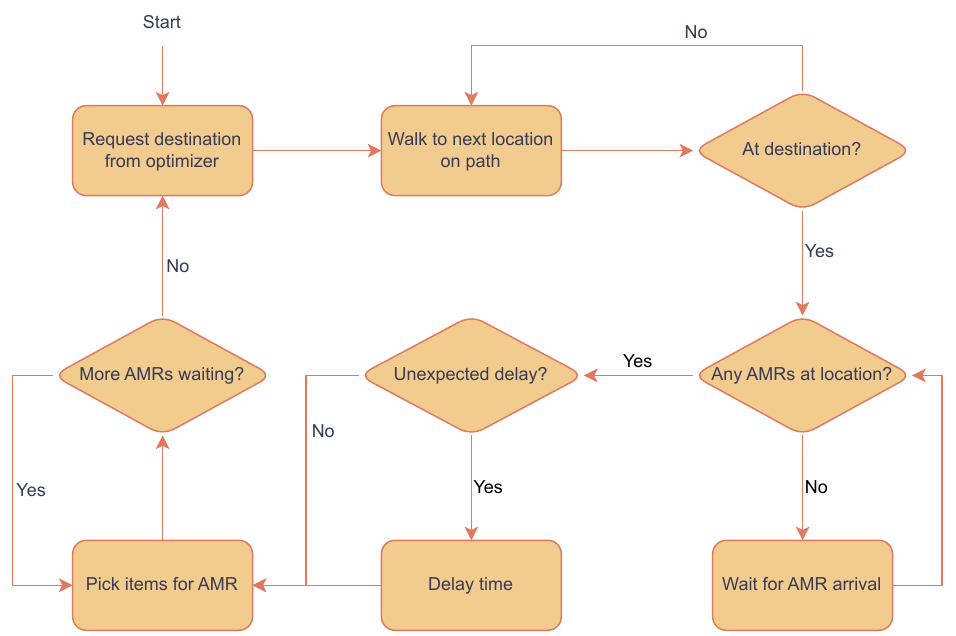}
    \caption{Overview of the picker process in the simulation model.}
    \label{fig: picker process}
\end{figure}

Figure \ref{fig: picker process} shows the schematic overview of the picker process in the simulation model.

\subsection{AMR Process}
\begin{figure}[ht]
    \centering
    \includegraphics[width=0.5\linewidth]{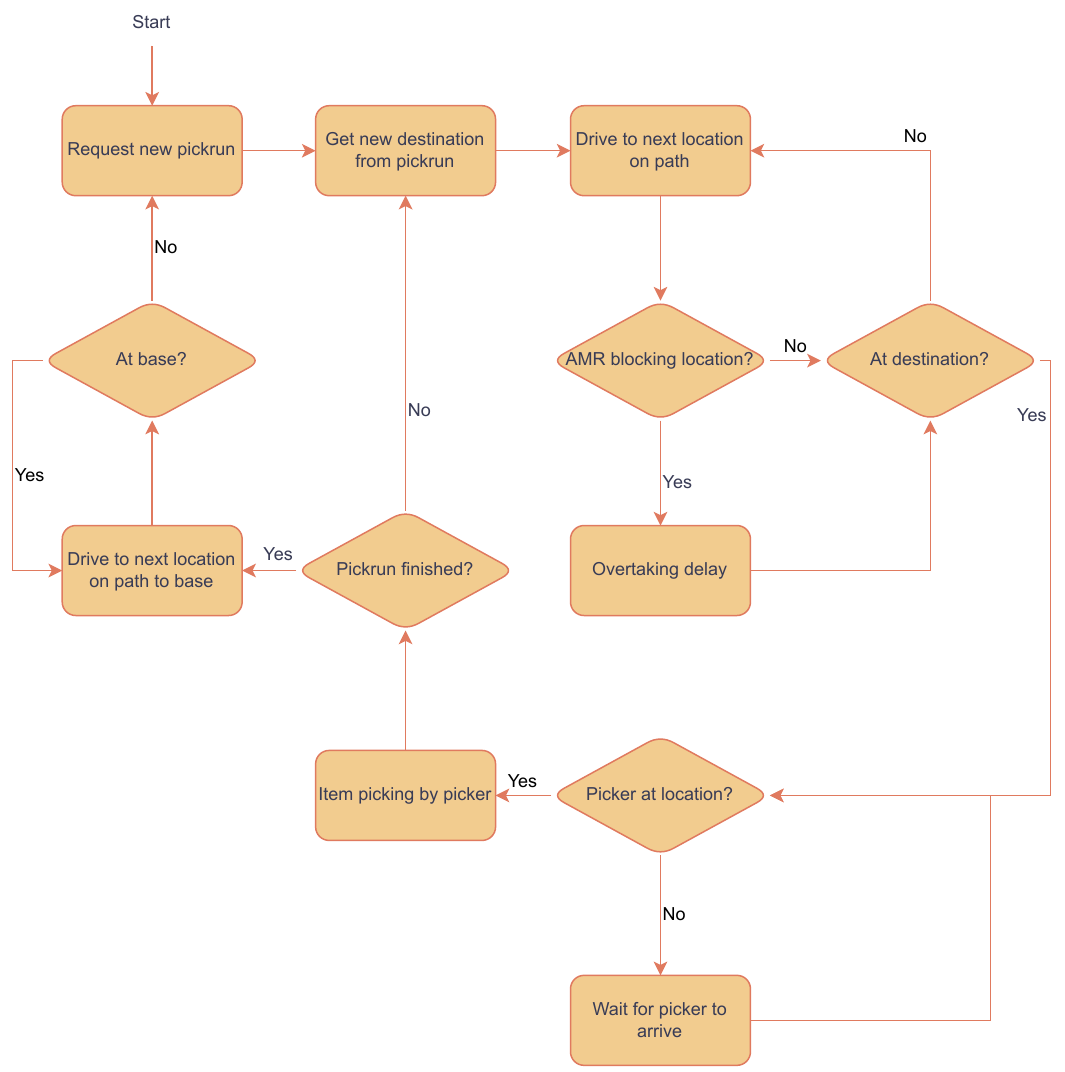}
    \caption{Overview of the \acrshort{amr} process in the simulation model.}
    \label{fig: amr process}
\end{figure}

Figure \ref{fig: amr process} shows the schematic overview of the \gls{amr} process in the simulation model.
